\documentclass{article}
\usepackage{arxiv}

\usepackage[utf8]{inputenc} 
\usepackage[T1]{fontenc}    
\usepackage{url}            
\usepackage{booktabs}       
\usepackage{amsfonts}       
\usepackage{hyperref}       
\usepackage{nicefrac}       
\usepackage{microtype}      
\usepackage{lipsum}		
\usepackage{graphicx}
\usepackage[numbers]{natbib}
\usepackage{doi}
\usepackage{multirow}

\usepackage[a-2b]{pdfx}
\usepackage{subcaption}
\usepackage[labelfont=bf]{caption} 
\usepackage{tabularx}

\usepackage{graphicx}
\usepackage{balance}
\usepackage{times}
\usepackage{tikz}
\usepackage{amsmath}
\usepackage{amsthm}
\usepackage{graphicx}
\usepackage{epstopdf}
\usepackage{latexsym}
\usepackage{algorithm}
\usepackage{algorithmic}
\usepackage{pifont}
\usepackage{epsfig}
\usepackage{stmaryrd}
\usepackage{array}
\usepackage[normalem]{ulem}

\usepackage{bm}
\usepackage{cleveref}
\usepackage{listings}
\usepackage{courier}
\usepackage{enumitem}
\usepackage{appendix}
\usepackage[frozencache,cachedir=minted-cache]{minted}
\usepackage{amssymb}
\usepackage{xcolor}

\usepackage[utf8]{inputenc}

\DeclareFixedFont{\ttb}{T1}{txtt}{bx}{n}{12} 
\DeclareFixedFont{\ttm}{T1}{txtt}{m}{n}{12}  

\usepackage{color}
\definecolor{deepblue}{rgb}{0,0,0.5}
\definecolor{deepred}{rgb}{0.6,0,0}
\definecolor{deepgreen}{rgb}{0,0.5,0}
\newcommand\pythonstyle{\lstset{
language=Python,
basicstyle=\ttm,
morekeywords={self},              
keywordstyle=\ttb\color{deepblue},
emph={MyClass,__init__},          
emphstyle=\ttb\color{deepred},    
stringstyle=\color{deepgreen},
frame=tb,                         
showstringspaces=false
}}

\lstnewenvironment{python}[1][]
{
\pythonstyle
\lstset{#1}
}
{}

\newcommand{\system}{{\sc Palimpzest}}

\usepackage{eqparbox}

\newcommand{\etal}{\textit{et al.}}

\newcommand{\circlednum}[1]{\raisebox{.5pt}{\textcircled{\raisebox{-.9pt} {#1}}}}

\lstdefinestyle{mystyle}{
    basicstyle=\ttfamily\footnotesize,
    breakatwhitespace=false,         
    breaklines=true,                 
    captionpos=b,                    
    keepspaces=true,                 
    showspaces=false,                
    showstringspaces=false,
    showtabs=false,                  
    tabsize=2
}

\ifdefined\finaltrue
    
    \newcommand{\st}[1]{}
\else
    
    \newcommand{\st}[1]{\textcolor{red}{\sout{#1}}}
\fi

\title{A Declarative System for Optimizing AI Workloads}

\author{Chunwei Liu\textsuperscript{*}, Matthew Russo\textsuperscript{*},  Michael Cafarella,\\ 
\textbf{Lei Cao\textsuperscript{\textdagger}, Peter Baille Chen, Zui Chen, Michael Franklin\textsuperscript{\ddag}},\\ 
\textbf{Tim Kraska, Samuel Madden, Gerardo Vitagliano} \\
\\
	MIT, \textsuperscript{\textdagger}University of Arizona, \textsuperscript{\ddag}University of Chicago\\
 \\
	\texttt{chunwei@mit.edu, mdrusso@mit.edu, michjc@csail.mit.edu}\\
 \texttt{caolei@arizona.edu, peterbc@mit.edu, chenz429@mit.edu, mjfranklin@uchicago.edu},\\ \texttt{kraska@mit.edu, madden@csail.mit.edu, gerarvit@csail.mit.edu} \\
 }

\date{} 					

\renewcommand{\shorttitle}{\textit{arXiv} preprint}

\begin{document}
\maketitle
\begin{abstract}
A long-standing goal of data management systems has been to build systems which can compute quantitative insights over large corpora of unstructured data in a cost-effective manner. Until recently, it was difficult and expensive to extract facts from company documents, data from scientific papers, or metrics from image and video corpora. Today's models can accomplish these tasks with high accuracy. However, a programmer who wants to answer a substantive AI-powered query must orchestrate large numbers of models, prompts, and data operations. For even a single query, the programmer has to make a vast number of decisions such as the choice of model, the right inference method, the most cost-effective inference hardware, the ideal prompt design, and so on. The optimal set of decisions can change as the query changes and as the rapidly-evolving technical landscape shifts. In this paper we present \system, a system that enables anyone to process AI-powered analytical queries simply by defining them in a declarative language. The system uses its cost optimization framework to implement the query plan with the best trade-offs between runtime, financial cost, and output data quality.  We describe the workload of AI-powered analytics tasks, the optimization methods that \system\ uses, and the prototype system itself. We evaluate \system\ on tasks in Legal Discovery, Real Estate Search, and Medical Schema Matching. We show that even our simple prototype offers a range of appealing plans, including one that is 3.3x faster and 2.9x cheaper than the baseline method, while also offering better data quality. With parallelism enabled, \system\ can produce plans with up to a 90.3x speedup at 9.1x lower cost relative to a single-threaded GPT-4 baseline, while obtaining an F1-score within 83.5\% of the baseline. These require no additional work by the user.
\end{abstract}


\keywords{Relational optimization \and LLMs \and AI programming}

\textsuperscript{*} indicates equal first author contribution.

\section{Introduction}
\label{sec:intro}





Advances in AI models have driven progress in applications such as question answering \cite{zhang2024endtoend}, chatbots \cite{vicuna2023}, autonomous agents \cite{patil2023gorilla,schick2023toolformer}, and code synthesis \cite{Li_2022,hong2023metagpt}. In many cases these systems have evolved far beyond posing a simple question to a chat model: they are  complex AI systems~\cite{compound-ai-blog} that combine elements of data processing, such as Retrieval Augmented Generation (RAG); ensembles of different models; multi-step chain-of-thought reasoning; and in many cases, cloud-based modules.

It is easy for the runtime, cost, and complexity of these AI systems to escalate quickly, particularly when applied to large collections of documents. Consider a few simple AI-powered analytical tasks:

\begin{itemize}
    \item {\bf Legal Discovery (\autoref{fig:legal-discovery-correct}}) --- In this use case, prosecutors conducting an investigation  wish to identify emails from defendants which are (a) related to corporate fraud (e.g., by mentioning a specific fraudulent investment vehicle) and (b) do not quote from a news article reporting on the business in question. Test (a) may be implemented using a regular expression or UDF, while (b) requires semantic understanding to distinguish between employees sharing news articles versus first-hand sources of information. An efficient implementation would recognize that (a) can likely be implemented using conventional and inexpensive methods, while (b) may require an LLM or newly-trained text model to retain good quality.
    \item {\bf Real Estate Search (\autoref{fig:real-estate-search-correct}}) --- In this use case, a homebuyer wants to use online real estate listing data to find a place that is (a) modern and attractive, and (b) within two miles of work. Test (a) is a semantic search task that possibly involves analyzing images, while (b) is a more traditional distance calculation over extracted geographic data.  Any implementation needs to process a large number of images and listings, limit its use of slow and expensive models, and still obtain high-quality results.
    \item {\bf Medical Schema Matching (\autoref{fig:biofabric-matching})} --- In the medical domain, cancer research is often based on large collections of information about patient cases and sample data. However, research studies' data outputs do not always conform to a unified standard. In this use case, based on the medical data pipeline described by Li, \etal~\cite{li2023proteogenomic}, we imagine a researcher who would like to (a) download the datasets associated with a dozen specified cancer research papers, (b) identify the datasets that contain patient experiment data, and (c) integrate those datasets into a single table.  Step (a) requires parsing and understanding research paper texts to obtain dataset URLs, step (b) requires classifying each dataset as either patient-related or not, and step (c) requires a complex data integration task. As with the use cases above, the programmer must manage multiple subtasks, each of which offer different possible optimization opportunities and quality trade-offs.
\end{itemize}

These tasks:
\begin{enumerate}
    \item Interleave traditional data processing with AI-like semantic reasoning
    \item Are data-intensive: each source dataset could reasonably range from hundreds to millions of records
    \item Can be decomposed into an execution tree of distinct operations over sets of data objects
    \item May result in answers of varying quality
\end{enumerate}

\noindent {\bf Semantic Analytics Applications:} Taken together, these criteria outline a broad class of AI programs that are important, complex, and potentially very optimizable; we call them {\em semantic analytics applications} --- or, \textbf{SAPPs}. We believe there is a large set of such use cases that mix conventional data analytics with transformations and filters that would not be possible without AI methods. Such workloads frequently require interleaved data acquisition steps, conventional analytical queries, and AI operations. The AI operations process unstructured data, require broad domain knowledge to implement, or have specifications that users may not be able to implement correctly with traditional source code. 

\noindent {\bf Challenges:} Naively scaling AI systems to process SAPPs with thousands or millions of inputs appears to require spending a huge amount of runtime and money executing high-end AI models. The performance gap between traditional data processing components and AI-powered components is profound.  For example, a high-quality open-source LLM running on a modern GPU might process 100-125 tokens per second. Assuming a token is represented by 5 bytes (on average), such a model yields a throughput of {\em less than 1 KB per second}. OpenAI's new GPT-4o model currently costs 5 USD for 1M input tokens, or in other words 5 USD for processing just 5MB of data. These numbers are many orders of magnitude worse than any other component of the modern data processing stack, such as data storage, network bandwidth, SQL query processing time, and so on.

Thus, optimizing the use of AI components is crucial, while at the same time current AI infrastructure is in a state of tremendous technical flux. New models and implementation techniques are published seemingly every day, while model costs and runtimes change constantly. Harnessing the latest advances in model runtime, cost, and quality is complex, error-prone, and requires engineers to constantly rewrite and retune their systems.

Consider the wide range of technical decisions an AI engineer faces:
\begin{itemize}
    \item When designing prompts, the engineer must optimize wording, choose to employ zero- or few-shot examples, and perhaps decide on a general prompting strategy (e.g., chain-of-thought, ReAct \cite{yao2023react}, etc.).
    \item When choosing models, the engineer must pick the best model {\em for each substask in the program}, balancing time, cost, and quality.
    \item When running the workload, the engineer must decide whether each subtask is best implemented by a foundation model query, synthesized code, or a locally-trained student model. Furthermore, they must consider how to combine tasks to improve GPU cache utilization, and how to avoid
    running over LLM context limits.
    \item When scaling out to a larger dataset, the engineer faces additional challenges in selecting an efficient execution plan. Even if the system performs well on a small dataset, it may require redesign to ensure reasonable runtime, cost, and performance at a larger scale. This may involve enabling parallelism for each component and integrating these parallelized components seamlessly into the broader system for optimal efficiency. 
    \item When integrating with external data systems, the engineer must decide how to choose parameters (e.g., the number of chunks to return per RAG query) in a manner that yields the best speed, cost, and quality trade-offs.
\end{itemize}

The space of possible decisions is vast, and choosing wisely depends on low-level details of the exact task being performed. Moreover, the definition of "best" can change over time: a developer might prefer "fast and cheap" execution when quickly testing initial proof-of-concept ideas, then switch to "costly but high-quality" for customer deployment. Finally, the changing technical landscape means that optimization choices made today might be obsolete tomorrow.

\begin{figure}[t!]
    \centering
    \includegraphics[width=0.85\textwidth]{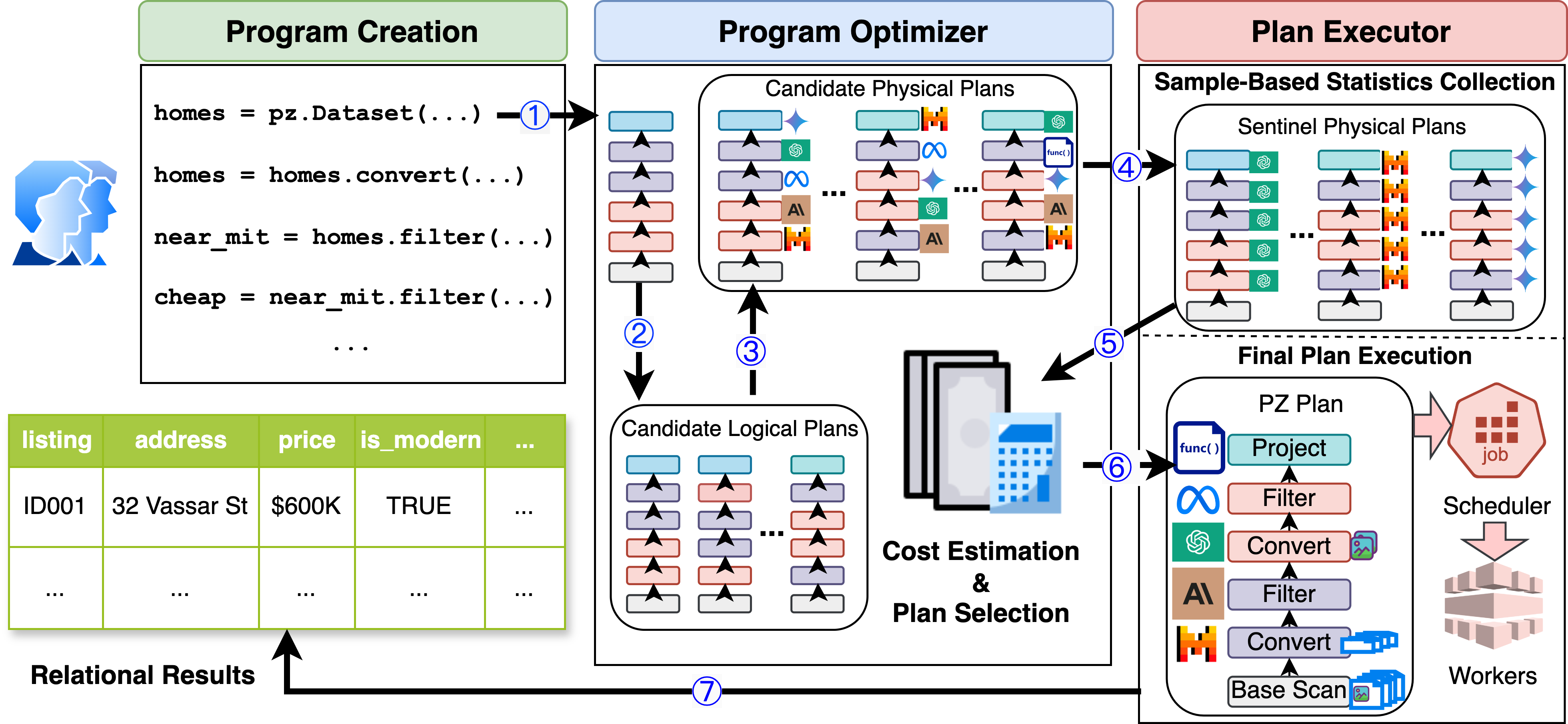}
    \caption{Overview of the \system{} system. Users write their program(s) in a declarative language which undergoes compilation \circlednum{1}, logical plan generation \circlednum{2}, and physical plan generation \circlednum{3}. Subsequent steps involve profiling sample plans \circlednum{4} and analyzing performance statistics to estimate costs \circlednum{5}. The optimal plan, tailored to user-specified preferences (e.g. to maximize quality at fixed cost), is selected \circlednum{6} and executed, delivering relational results to the user \circlednum{7}. This comprehensive process is designed to optimize execution by effectively balancing cost, runtime, and quality.
    }
    
    \label{fig:intro}
\end{figure}

\noindent {\bf Our Goal:} The key insight is that machines, not human engineers, should decide how best to optimize semantic analytics applications. Engineers should be able to write AI programs at a high level of abstraction and rely on the computer to find an optimized implementation that best fits their use case. A similar set of circumstances --- a need for performance improvements for an important workload, during a time of enormous technical change --- led to the development of the relational database query optimizer in the 1970s. Today's underlying technical challenges are very different, but the basic idea of declarative program optimization remains valuable.

In this paper we lay out our vision and a prototype for \system\footnote{Like an ancient palimpsest, our system entails constant revision and rethinking, in our case by the optimizer. Only zestier!}, a system that enables engineers to write succinct, declarative code that can be compiled into optimized programs. \system{} is designed to optimize the broad SAPP workload class, which should encompass large-scale information extraction, data integration, discovery from scientific papers,  image understanding tasks, and multimodal analytics. As shown in \autoref{fig:intro}, when running an input user program, \system\ considers a range of logical and physical optimizations, then yields a set of possible concrete executable programs. \system\ estimates the cost, time, and quality of each one, then chooses a program based on runtime user preferences. The system is designed to be extensible, so that new optimizations can be easily added in the future. Just as the RDBMS allowed users to write database queries more quickly and correctly than they could by writing traditional code, \system\ will allow engineers to write better AI programs more quickly than they could unaided.

\noindent {\bf Our Approach:} A core challenge in building \system\ is creating an optimizer that can marshal many optimizations to meet a user's cost, runtime, and quality goals. By using a language that is high-level, type-focused, and declarative --- rather than the low-level prompting and coding method pursued by naive programming and some other frameworks \cite{langchain, zheng2023efficiently} --- we believe \system\ can exploit many optimizations that are not otherwise available. Another key challenge involves designing a programming interface which simultaneously enables engineers to express the broadest possible set of AI programs, while imposing structure on their programs that the optimizer can exploit. To this end, we created a Python library which implements a thin abstraction over an underlying relational algebra. The core intellectual difference between \system\ and previous database-style systems is the addition of the relational {\bf convert} operator, which transforms an object of one user-defined schema to another. This operator --- which is implemented using a variety of methods, often based on foundation models --- allows the programmer to implement many AI tasks in a relational and optimizable style.

\noindent {\bf Contributions:} In this paper we:
\begin{itemize}
    \item Introduce Semantic Analytics Applications (SAPPs), a new class of data-intensive AI workloads that can benefit from many traditional ideas in data management. Addressing them requires a range of new technical solutions and abstractions. (\autoref{sec:workloads}.)
    \item Describe the \system\ architecture and how it aims to support SAPPs. (\autoref{sec:overview}.)
    \item Describe a set of physical and logical optimizations, several of which are implemented in our prototype. (\autoref{sec:optimizations}.)
    \item Present experimental results that show that even with just these basic optimizations in place, \system\ can execute SAPP workloads with a range of tradeoffs that are more appealing than a baseline approach.  The exact benefit depends on the workload and whether the user prefers to reduce time, reduce cost, or maximize quality. Our results include, for example, a plan that is 3.3x faster, 2.9x cheaper, and offers better data quality than its baseline method; and another plan that is 4.7x faster and 9.1x cheaper, with a trade-off of 14.3\% lower quality than its baseline. (\autoref{sec:evaluation}.)
    \item Demonstrate that our prototype can produce parallelized plan implementations which are up to 90.3x faster and 9.1x cheaper than a single-threaded GPT-4 baseline, with an F1-score within 83.5\% of the baseline. (\autoref{sec:evaluation}.)
\end{itemize}


\system\ is an exciting prototype system and we are looking forward to the potential for future optimizations and features. We invite interested readers to experiment with our code at \url{https://github.com/mitdbg/palimpzest} and perhaps contribute some source code of their own.

\section{Workloads}
\label{sec:workloads}
\begin{figure*}[t!]
    \begin{subfigure}[t]{0.48\textwidth}
        \includegraphics[width=\textwidth]{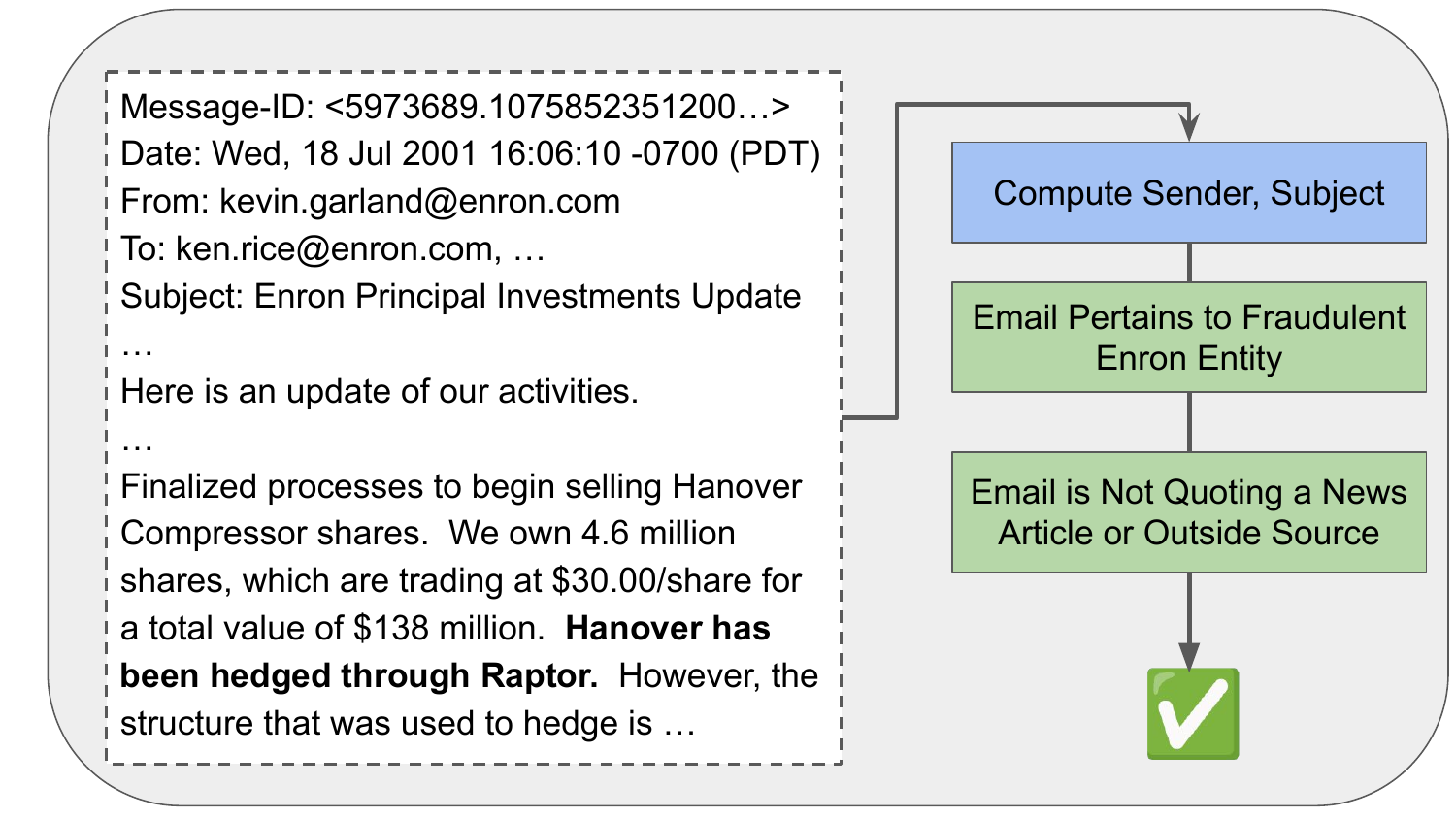}
        \caption{Example of a positive entry in the Legal Discovery workload. This email meets the criteria of (1) mentioning a fraudulent entity (in this case, ``Raptor") and (2) not quoting from a news article or a source outside of Enron.}
    \label{fig:legal-discovery-correct}
    \end{subfigure}
    \hfill
    \begin{subfigure}[t]{0.48\textwidth}
        \includegraphics[width=\textwidth]{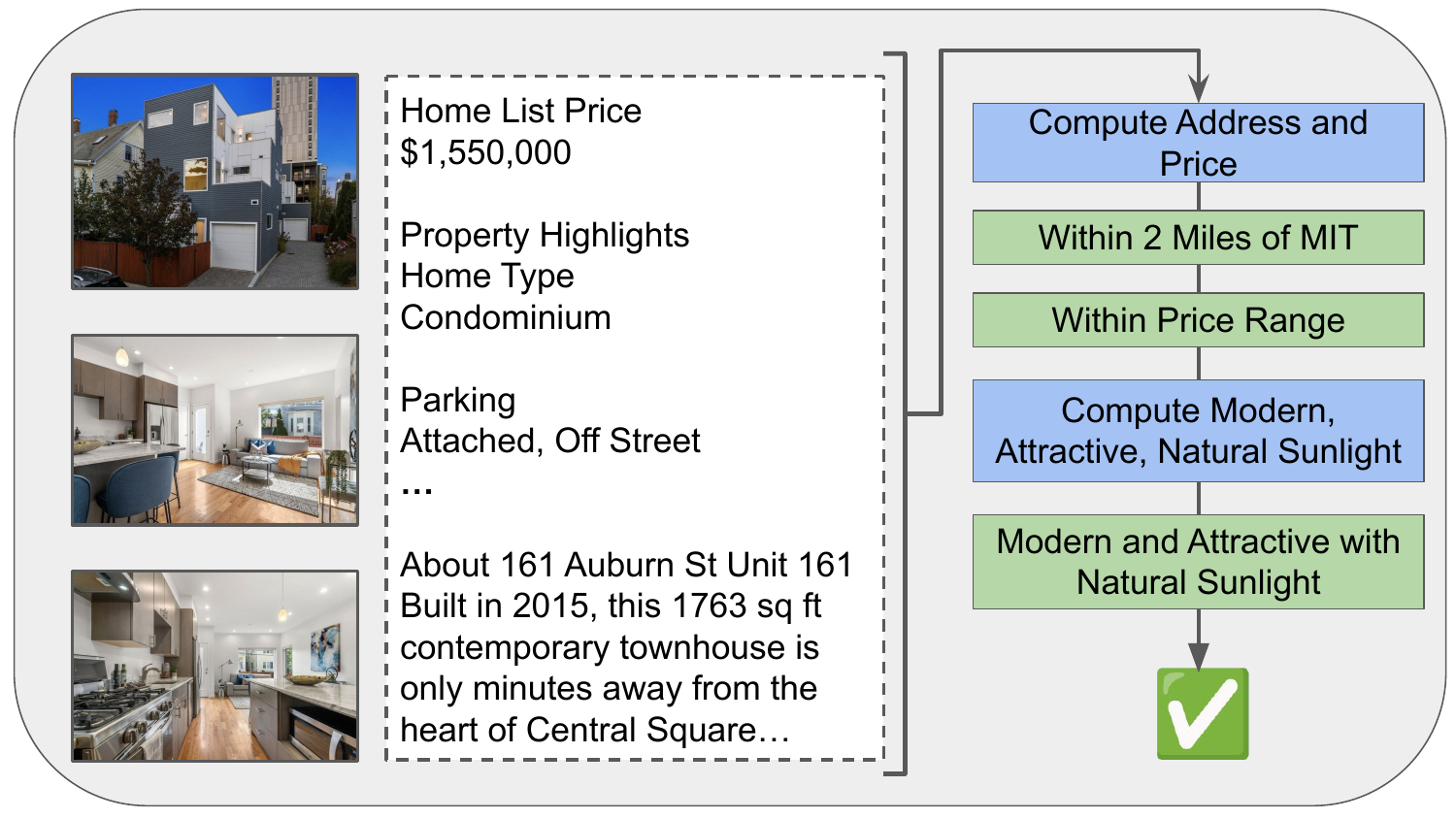}
        \caption{Example of a positive entry in the Real Estate Search workload. The house meets the three criteria of being (1) close to MIT (2) within the user's price range and (3) modern and attractive with lots of natural sunlight.}
    \label{fig:real-estate-search-correct}
    \end{subfigure}
    \centering
    \begin{subfigure}[b]{0.95\textwidth}
        \includegraphics[width=\textwidth]{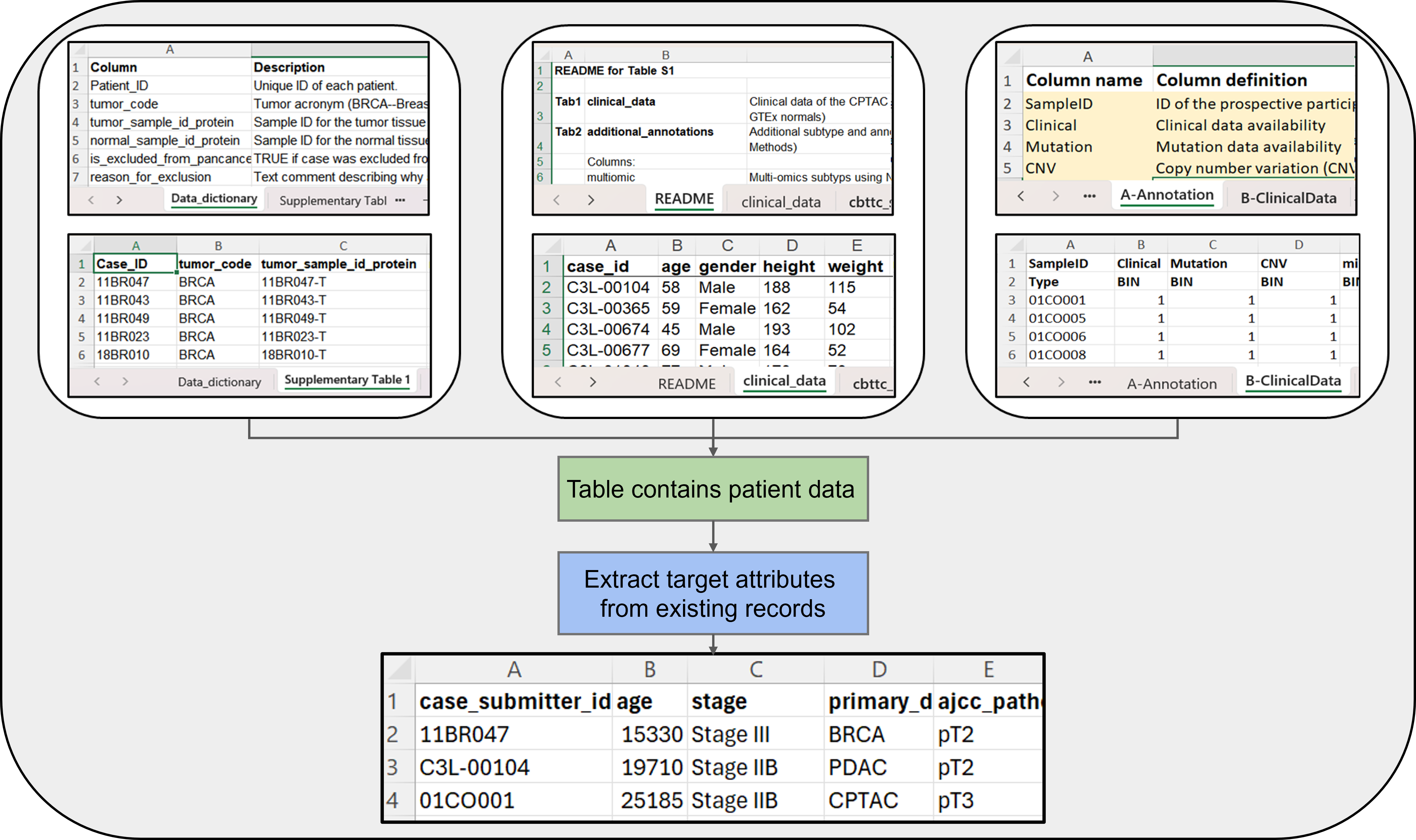}
        \caption{Example of the Medical Schema Matching workload. Three spreadsheets contain several tables and \system{} (1) filters for tables that relate to patient data and (2) matches and extracts the relevant attributes, consolidating a single table with a target schema.}
        \label{fig:biofabric-matching}
    \end{subfigure}
    \caption{Positive examples from the Legal Discovery and Real Estate Search workloads, as well as an overview of the Medical Schema Matching workload. Negative examples can be found in the Appendix (\autoref{fig:negative-examples}).}
    \label{fig:workloads}
\end{figure*}

Before we describe the details of the \system{} system, it is useful to discuss the workloads \system{} aims to support, in particular the SAPP workloads.

As discussed in \autoref{sec:intro}, SAPPs (1) combine traditional data processing and AI elements, (2) potentially process large amounts of data, and (3) can be decomposed into a tree of operations over sets of data objects. As a running example, consider the Real Estate Search task in \autoref{fig:real-estate-search-correct}. In this task, the user wants to search all of the real estate listings near Cambridge, MA to find a house that is (a) modern and attractive, (b) within two miles of MIT, and (c) within the user's price range.

This task clearly satisfies the first criterion: determining whether a house is ``modern and attractive" likely requires using some vision or text model to process the textual listings and images. Moreover, depending on the dataset, recovering price and location data may require extraction from text.  The second criterion is satisfied because --- as of this writing --- Zillow shows 1,327 listings for homes in the Cambridge and Boston areas, and each listing contains a text description along with (typically) 10 or more images. As for the third criterion, the query clearly has a set of objects as both input and output. It is easy to see how we might first have an operator that extracts price and location fields from the raw data, then a second that applies traditional selection predicates. Finally, the system might have an operator that extracts whether the images show a "modern and attractive" residence.  \autoref{fig:real-estate-search-correct} shows a conceptual pipeline, with imperative steps listed in blue, satisfied logical tests in green, and failed logical tests in red.

{\bf Optimization Challenges.} The ideal implementation of an AI system for a SAPP workload will jointly optimize its AI- and conventional data processing elements. For example, in Real Estate Search, an extremely naive implementation might waste time and money processing images of apartments to test whether they are ``modern and attractive" --- only to discard them when they fail to meet conventional constraints. A slightly more sophisticated implementation would reorder the plan so that it applies the restrictive conventional constraints first, thereby avoiding the time and expense of invoking a vision model on candidates that will later be discarded. Further optimizations might process the text description first to evaluate whether an apartment is likely to be ``modern and attractive", and then apply an image processing method only when deemed necessary. The decisions around which optimizations to use and when to use them will have to be made for each new task and dataset.


At its core, optimizing an AI system to execute a given SAPP workload requires making accurate predictions about the runtime, cost, and quality of each semantic and conventional data processing step. Estimating these metrics for semantic tasks can be particularly challenging. For example, estimating the runtime and cost for a vision model requires knowing the average number of input and output tokens per record as well as the total number of records that will be processed by the model (i.e., its cardinality). Estimating the quality of an output --- especially without labelled data --- may require using heuristics that can be error prone, or comparing against an expensive "champion" model. Finally, for every physical optimization available to the system (such as using an ensemble of vision models or decreasing the image resolution), the optimizer needs to predict the optimization's impact on these metrics.



{\bf System Design Challenges.} An ideal system for processing SAPP workloads will also require minimal engineering effort to maintain application code over time, in the face of changing user needs and input data. To the extent that AI components have been integrated into data applications in the past, significant effort went into training models narrowly tailored for a particular semantic task. Outside of a few tech giants, models were infrequently retrained. Models were restrictive enough that the space of system-wide performance tuning decisions was not large, and at any rate models changed infrequently enough that system-wide tuning did not have to happen often.

The modern AI landscape has changed all of these assumptions. The set of models and adjacent AI approaches (such as prompting strategy) is large and rapidly changing. The space of system-wide performance decisions is vast, and the best decision changes with each new model, task, business need, and dataset. Anecdotally, we have heard several RAG users and vendors start to complain that deploying these systems requires making repeated configuration decisions that reflect reasonable performance vs quality trade-offs; this is a task that traditional data administrators are often not equipped to make. \system{} aims to enable AI system engineers to focus on programming the system logic while letting the optimizer select which models and physical optimizations are needed to meet the user's preferences for cost, runtime, or data quality.

\section{Overview}
\label{sec:overview}
We present an overview of the \system{} system and discuss its primary components. 
First, we describe \system{}'s relational model, the convert operator, and how AI workloads can be interpreted as computing relational views. Then we discuss a short sample program and language features.  Finally, we provide a high-level view of the system's architecture, including walking through a simple program execution.

\subsection{Semantic Analytics and the Relational Model}
\begin{figure}
  \centering 
  \begin{minted}[xleftmargin=17pt,linenos,escapeinside=||,fontsize={\fontsize{8.5}{7.5}\selectfont}]{python}
import palimpzest as pz

class Email(|\textcolor{blue}{pz.TextFile}|):
  """Represents an email, which can subclass a text file"""
  sender = |\textcolor{blue}{pz.StringField}|(desc="The email address of the sender", required=True)
  subject = |\textcolor{blue}{pz.StringField}|(desc="The subject of the email", required=True)

# define logical plan
emails = |\textcolor{blue}{pz.Dataset}|(source="enron-emails", schema=|\textcolor{blue}{Email}|) # invokes a convert operation
emails = emails.filter("The email is not quoting from a news article or an article ...")
emails = emails.filter("The email refers to a fraudulent scheme (i.e., \"Raptor\", ...")

# user specified policy
policy = |\textcolor{blue}{pz.MinimizeCostAtFixedQuality}|(min_quality=0.8)

# execute plan
results = |\textcolor{blue}{pz.Execute}|(emails, policy=policy)
    \end{minted}
  \caption{The AI program written using \system{} for the Legal Discovery workload.}
  \label{fig:enron-email-program}
\end{figure}
    




\system\ treats its programs primarily as a form of computing {\em relational views}: the user specifies a (set of) input relation(s) (called {\tt Datasets}) and a target output relation to be computed. Each relation has a corresponding {\tt Schema}. The user also describes a series of operations to be applied that transform the inputs into the output.

\autoref{fig:enron-email-program} shows a short example program, which we used for our evaluation in \autoref{sec:evaluation}. In this program, the user wants to identify emails that are not quoting from sources outside of Enron and that reference fraudulent investment vehicles. As a first step, the programmer uses \system{} to create a custom schema for the input dataset of Emails -- on lines 3-6.
In this case, Email is a subclass of {\tt TextFile}, which is defined in \system{}'s core library and inherits directly from the base {\tt Schema} class.  Starting on line 9, the user begins to describe data processing actions, beginning with instantiating an initial {\tt Dataset} that adheres to the {\tt Email} schema. The {\tt source} string ``enron-emails" uniquely identifies a set of files that have been preregistered with \system\ (see \autoref{sec:registration} for more details). The code on line 9 transforms the raw input data objects into the {\tt Email} Schema and stores the results int the {\tt emails} Dataset. On line 10, the program filters {\tt emails} for the subset which are not quoting from news articles. On line 11 the program filters for emails which discuss fraudulent investment entities.

The programmer takes two more steps: on line 14, she specifies a {\tt policy} that describes how the system should choose among multiple possible implementations of the steps described so far. (In this case, the plan with the lowest expected financial cost, subject to a lower bound on quality, is preferred.) Finally, on line 17, the programmer asks \system\ to {\tt Execute()} the program; this entails generating a logical execution plan, generating multiple optimized physical execution plan candidates, choosing one according to the specified policy, and then executing the code and yielding results. Programs written with \system{} are executed lazily, thus no actual data processing occurs until line 17.

The user-provided description strings for the schema fields and filters comprise both a way for the developer to specify correct program output, and a way for the system to find a high-quality implementation. As we will discuss more in \autoref{sec:convert} and \autoref{sec:quality}, these strings are provided to underlying operators which use them when constructing internal prompts. In contrast to prompt engineering, we do not intend for users to expend significant effort tuning these descriptions. (In our own evaluation, we set our field and filter descriptions once and never modified them). Instead, \system{} tunes the prompts automatically for the user.


Unlike SQL, \system\ is intended mainly to be used as a library in a host language; although the current implementation is in Python, there is no reliance on unique features of the Python language,
and porting the system to other languages would be straightforward. Even a host-independent program syntax, akin to SQL, would be possible. However, since we view easy programmatic integration with other data processing elements as a core design goal, we think that a strong hosted-language system is a better design choice.

\autoref{fig:algebra} shows the logical relational operators supported by \system{}.  Some of the operators, such as groupby, aggregation, and limit, are not showcased in \autoref{fig:enron-email-program}. These operators currently follow their standard definitions from the data management literature, but in the future groupby and aggregation could also be implemented using AI-based operations. For example, an AI could be asked to group documents based on sentiment, or to compute the largest dog in a set of images. Line 9 --- which creates the initial {\tt Dataset} --- implies a new operator: the \textbf{Convert} operation. This operator transforms a record of one schema into a record of another schema. In this case, the source data objects have a default schema of {\tt File}, which must be converted to examples of {\tt Email}. This operator is how \system\ implements most of its AI-intensive steps and is worth describing in detail.

\subsection{Convert}
\label{sec:convert}
Convert transforms a typed data object into a new object with a specified schema. Specifically, given an input object of Schema A and an output object of Schema B, the convert operation will produce the set of fields in Schema B which do not already exist in Schema A. One benefit of this design is that schemas can be defined incrementally. For example, if a user has defined a base schema \texttt{Animal} with fields such as \texttt{name}, \texttt{kingdom}, and \texttt{description}, they could then define schemas for individual animals such as \texttt{Dog}. The new class would inherit from \texttt{Animal} and only specify fields the user wishes to compute for \texttt{Dogs}. If the convert operation uses an LLM for its physical implementation, the fields of Schema A will be marshaled into a prompt as key-value pairs (along with the user-provided field descriptions) and the LLM will be asked to produce the output field(s) for Schema B.


A point of emphasis is that {\bf the user does not need to specify how to implement a particular convert operation}.  Instead, it is the system's job to implement and perform the operation. By employing a range of different AI models and generation techniques, \system\ can automatically compute the conversion function for each pair of schemas observed in the user's program. This includes using built-in operators for some base type conversions (e.g. \texttt{File -> TextFile} and \texttt{XLSFile -> Table}). In some cases, users may wish to hard-code convert operations using lambda expressions for guaranteed conversions, but this is not required.




Executing a particular convert operator may entail textual information extraction, text summarization, classification, image understanding, or any number of other AI tasks. The correct behavior of a convert operation is in most cases implied only by the user's specification of the input and output {\tt Schemas}. In the case of Figure~\ref{fig:enron-email-program}, the convert operator must extract the {\tt sender} and {\tt subject} fields from a raw data object in order to produce an {\tt Email}.


Here are a few examples of useful convert tasks, from simple to more complex:
\begin{itemize}
  \item The system can convert a {\tt TextFile} into an {\tt Email} by extracting the email's sender and subject (\autoref{fig:enron-email-program}).
  \item It can convert a {\tt PDFFile} into a {\tt ScientificPaper} by extracting the title, authors, abstract, and citations if the content includes elements of a scientific paper, returning None otherwise. (\autoref{fig:examples}).
  \item It can convert an {\tt Image} into a {\tt DogBreed} by extracting the breed of any dogs that are present and returning None otherwise. (\autoref{fig:examples}).
  \item The system can convert an {\tt Email} into a {\tt SlackMessage} by computing a summary of the email's contents. (\autoref{fig:examples}).
\end{itemize}


In some cases, the user might specify a convert task that the system cannot figure out how to perform, either because the operator is not implemented well, or because the user's conversion request makes no sense. In such cases, the current prototype will always drop records for which the conversion fails and continue processing the input {\tt Dataset}. In the future we intend to support additional behaviors, such as logging warning messages or aborting the program. When converting a Dataset of Schema A to Schema B, the resulting set may contain fewer records than the original (e.g., if some records are dropped); it may contain a one-to-one mapping of records; and it may also contain more records than the original (e.g., if the conversion extracts each author of a scientific paper into a separate record or more than one dog from an image).

We think the convert operator will be especially useful when the input object comprises multiple low-level datatypes, such as a real estate listing that has information in both image and text. Many listings will contain price data in text, but some include price directly embedded inside the images.  Ideally, users should not have to meticulously specify which specific real estate subfield contains the essential information; instead, the system will simply figure it out.  At the moment, implementing this flexibility is future work.


The implementation of the convert operator is intentionally abstracted from the user, as its internal mechanics can vary between executions and may not be representable in a standard transformation language due to the diverse input and output schemas it handles. This operator accepts data in one schema and outputs it in another, adapting dynamically to changing execution conditions.  While our current system often uses LLM inference to implement convert, the user cannot count on it. However, implementation details might be accessible through administrator-style tools.

The convert operator is meant to transform
data instances --- not models --- but it has similarities to some concepts from the model management literature. It is close to an execution of the
{\sc ComplexMatch} operator described by Bernstein, followed by an execution of the resulting mapping
on a particular data record~\cite{bernstein03}.  A single  convert execution is akin
to writing and running an entire program in the Potter's Wheel language~\cite{hellerstein}. 


In some cases, the user's program
may be ambiguous enough that several different convert implementations are arguably correct. This brings us to a core issue when designing \system: how to reason about quality and correctness.

\subsection{Correctness and Quality}
\label{sec:quality}
A significant design challenge for \system\ lies in how to specify correctness and manage quality.

Our design goal is to allow the user to treat data quality like she does performance in a modern RDBMS: It is usually good enough with no effort at all; but in rare cases where the quality is not good enough, improving through extra effort is possible. This philosophy extends to both the specification of correctness goals in the program and the practical achievement of high-quality data outputs at runtime.

\noindent {\bf Correctness Goals:} During specification --- that is, when the user is writing the program --- there is ideally enough information in the source and destination schemas to fully indicate the desired AI steps. But what if the {\tt ScientificPaper} Schema in \autoref{fig:examples} had the field {\tt institution}?  Should the system populate it with the institution of the first author, or the most-frequently-seen institution among the others, or the institution associated with the publisher?  

In most cases we expect a description of the goal, via field names and a simple text description, will be enough. However, if needed, our prototype provides users two primary levers for improving program correctness. First, the user can modify their logical plan's field and filter descriptions to be more precise. These descriptions are used to construct prompts for the convert and filter operations (either for direct invocation of an LLM or to help synthesize code), thus improving these descriptions can improve program correctness. Second, the user can rewrite their logical plan to have more concretely defined operations. For example, rather than writing our logical plan in \autoref{fig:enron-email-program} with a single filter to identify fraud-related emails, we broke the filter into two which were more well-defined.

We are working on support for the user to provide validation examples in the program text.

\noindent {\bf Output Quality:} A related concern is improving the system's ability to deliver high-quality outputs once the user's specification is clear. There are two subproblems. The first subproblem lies in accurately assessing the quality of a novel plan. The second subproblem lies in finding approaches that maximize assessed data quality.

In our prototype, \system{} assesses quality of a novel plan by 
using a ``champion model" approach, in which a powerful model (e.g., GPT-4) is used as a stand-in for ground truth against which all other methods' outputs are evaluated. In the near future, we will allow the user to provide labeled examples to compare against a plan's output.

Improving output quality permits a range of approaches.  Currently we rely entirely on different optimization approaches, such as model selection and code synthesis, to obtain high-quality output.  Once again, allowing the user to provide labeled examples will let us create a few-shot LLM prompt, or in some cases, employ reinforcement learning from human feedback \cite{ouyang2022training, rafailov2023direct}) methods.  We will also explore DSPy-style prompt modification.

\begin{figure*}[t!]
    \begin{subfigure}[t]{0.55\textwidth}
        \begin{minted}[escapeinside=||,fontsize={\fontsize{8.5}{7.5}\selectfont}]{python}
import palimpzest as pz

class ScientificPaper(|\textcolor{blue}{pz.PDFFile}|):
  """Represents a scientific research paper"""
  title = |\textcolor{blue}{pz.StringField}|(desc="Title of the paper")
  authors = |\textcolor{blue}{pz.ListField}|(desc="Author names", ...)
  abstract = |\textcolor{blue}{pz.StringField}|(desc="Paper abstract")
  citation = |\textcolor{blue}{pz.StringField}|(desc="Paper citation")

# load pdfs, convert, and filter
pdfs = pz.Dataset(source="papers")
papers = pdfs.convert(schema=ScientificPaper)

###############################################
class DogBreed(|\textcolor{blue}{pz.ImageFile}|):
  breed = |\textcolor{blue}{pz.StringField}|(desc="The dog's breed")

# load images, convert, and filter
images = pz.Dataset(source="my-dog-pics")
breeds = images.convert(schema=DogBreed)

###############################################
class SlackMessage(|\textcolor{blue}{pz.Schema}|):
  short_msg = |\textcolor{blue}{pz.StringField}|(desc=
    "A short summary message to be sent in Slack."
  )

# load emails, convert, and filter
emails = pz.Dataset(source="emails")
msgs = emails.convert(schema=SlackMessage)
        \end{minted}
        \caption{Example showing three AI programs. In the first, a dataset of PDF files is converted into {\tt ScientificPaper}. In the second, an image dataset is converted into {\tt DogBreeds} if the image contains a dog. In the third, emails are summarized into short Slack messages.}
        \label{fig:examples}
    \end{subfigure}
    \hfill
    \begin{subfigure}[t]{0.4\textwidth}
        \vspace{0pt}
        \renewcommand{\arraystretch}{1.3}
        \begin{tabular}{|c|l|}
        \hline
        {\bf operator } & {\bf description } \\
        \hline
        Project & $\pi$(rel., cols) \\
        \hline
        Select & $\sigma$(rel., predicate) \\
        \hline
        Convert & $\chi$(rel., schema\_a, schema\_b) \\
        \hline
        Group By & $\Gamma$(rel., group\_cond., agg.) \\
        \hline
        Limit & $L$(rel., limit) \\
        \hline
        Agg. & $\alpha$(rel., agg\_func) \\
        \hline
        \end{tabular}
        \caption{\system{}'s full relational algebra. We extend the traditional relational algebra to include operators such as groupby which produce multiple relations.}
        \label{fig:algebra}
    \end{subfigure}
    \caption{\system{} code examples and a summary of its full relational algebra.}
\end{figure*}

\subsection{Cost Optimization Framework}

\label{sec:planning}

At its core, \system{} allows users to define and execute logical \emph{plans}, which are sequences of relational operations on datasets. By design, the declarative nature of these plans leaves many details of \textit{how} to execute them underspecified. The key role of \system's cost optimizer is to identify physical implementations of these plans which are (near) optimal and align with user-specified preferences. The system diagram showing the path from program implementation, to generating and selecting the most cost-effective plan, and finally to executing that plan is shown in \autoref{fig:intro}. We reference steps in the diagram by number (e.g. step \circlednum{1}).  The details of this optimization process are given in Section~\ref{sec:optimizations}, but we briefly summarize the process here.

Developers first write declarative programs, such as the one shown in \autoref{fig:enron-email-program}. In this program, the user has specified a chain of Datasets and processing steps, which culminate in the final {\tt emails} set. Upon calling {\tt Execute()} on line 17, that chain is sent to the {\bf Program Optimizer} for compilation into an initial logical plan (step \circlednum{1}). The set of logical operators is shown in \autoref{fig:algebra}. Given this initial logical plan, \system{} performs logical optimization to create a set of new, functionally equivalent plans which may have different cost, runtime, and quality trade-offs (step \circlednum{2}). Many of these optimizations reuse traditional ideas from query optimization in databases, such as predicate pushdown, projection pushdown, and filter reordering. However, implementing these optimizations in \system{} presents new challenges. For example, effectively implementing projection pushdown may require determining the best way to split a single convert operation into multiple operations. 

The resulting logical plans are then used to generate an even larger set of candidate physical plans (step \circlednum{3}). This is akin to the physical optimization stage in relational databases. At this stage, the system hypothesizes a large number of concrete plans that are consistent with the input logical plans, but which make a range of detailed decisions specific to AI systems. These include decisions around model choice, k-v cache management, prompt generation, token reduction, and other areas. One LLM-specific optimization we implement is to combine multiple subtasks into a single LLM query (such as asking for both the {\tt title} and {\tt authors} of a scientific paper) in order to avoid unnecessary token duplication across tasks. This is similar to FrugalGPT's {\em query concatenation} method~\cite{chen2023frugalgpt}, but works at the compiled subtask level.

Note that much of the flexibility of the \system\ optimization process comes from the wide number of possible implementations for even one particular convert operation. For example, converting a {\tt TextFile} to an {\tt Email} could be done by formulating an LLM task or by synthesizing appropriate traditional code. Converting a {\tt PDFFile} to a {\tt ScientificPaper} could be done by using a set of LLM tasks or perhaps by training a conventional local NLP model. LLM tasks can be modified by choosing a different model, by changing the prompt, by combining multiple tasks into a single prompt, or in some cases even by first fine-tuning a model. Each of these options presents different trade-offs. Of course, in some cases users may wish to explicitly state how to perform some operations, in which case they are free to provide UDFs --- but this is not required.

At this point, the Program Optimizer has created a potentially large set of programs that are consistent with the user's input program and which operate at different points in the optimization space of runtime, financial cost, and quality. However, \system{} still needs to compute estimates of these metrics for each physical plan (steps \circlednum{4} and \circlednum{5}). To accomplish this, the {\bf Plan Executor} executes a small set of {\bf sentinel plans} to gather sample data on plan execution statistics. The quality of plan outputs are then evaluated against the output from a ``champion" plan as described in \autoref{sec:quality}, at the granularity of an individual operator. (We currently test against the plan which uses GPT-4 for every operation). This way, the system can score how well each model performs on each operation in a given plan. We provide more details on the specifics of this quality estimation procedure in \autoref{sec:choose-opt}. The Program Optimizer then chooses the best physical plan, according to per-plan estimates and the user's preference (step \circlednum{6}). Finally, this choice is sent back to the Plan Executor, which executes the plan and spends computation and financial resources, possibly drawing on many external model and data service providers (step \circlednum{7}).





\subsection{Dataset Registration and Result Caching}
\label{sec:registration}
Users must preregister named base-level source \texttt{Datasets} --- like {\tt enron-emails} in \autoref{fig:enron-email-program} --- before they can be consumed by \system{} operators at runtime. Giving every dataset an unambiguous name is meant to eventually allow an AI program to be run in different locations on the exact same set and thereby yield identical results. (However, 
the current prototype supports only local names. We will support a global dataset naming service in the future.)

The \system{} standard library implements a number of common datasource types already, including loading data from a single file or a single directory. In the future we will support relational databases, AWS S3, and other data sources. Users can also easily add custom datasources for their specific use cases. 

Given the low-throughput associated with invoking LLM models and services, \system{} makes a best-effort attempt to avoid sending requests to an LLM unless it is absolutely necessary. As a result, caching is a key component of the \system{} system.  The system currently  caches intermediate results at the granularity of a \texttt{Dataset}, but in the future we may modify this to cache individual records.  Thanks to unambiguous base-level \texttt{Dataset} names, \system\ can detect when two different programs share computation, and thus when a cached intermediate result from program A can be re-used by program B.

A caveat with caching is that different physical plans can yield results of different quality, so two runs of the same program may yield different results. In the future, the caching layer will record quality statistics with each cached intermediate result, so the system can invalidate previously cached results when user quality preferences change.

\section{Program Optimization}
\label{sec:optimizations}
Managing and exploiting a large space of useful optimizations is \system's core feature. In this section we describe key logical and physical optimizations that can be used by \system{} --- many of which have already been implemented in our prototype. 

\subsection{Logical Optimizations}
\label{sec:logical_opt}
\system{} has a set of logical optimizations which can be applied to the logical plan implied by a user's program. These optimizations create new plans that are logically equivalent, but may yield significant runtime and cost savings depending on factors like operation execution cost and the selectivity of filter operations.  We have implemented two logical optimizations already --- filter reordering and convert reordering --- but we plan on adding more in the future (such as splitting and coalescing LLM requests).

{\bf Filter Reordering} simply permutes the ordering of selection filters in a logical plan wherever possible. For example, if an input logical plan has the filter sequence A, B, and C, this logical optimization would yield 5 additional plans, one for each unique permutation of filters. If each of the filters has a unique selectivity, then exactly one of these permutations will minimize the number of records processed by the system. \system\ does not initially know these selectivities (though future versions may use historical data). However, \system{}'s cost optimizer is capable of estimating these selectivities after execution begins and sample execution data is collected. Thus, by considering all filter reorderings in its logical optimization stage, \system{} can find more efficient executions of user plans. 


    
\noindent {\bf Convert Reordering} enables \system{} to move convert operations around the logical plan, similar to filter reordering. This can be especially beneficial when a plan has a mixture of convert operations and filter operations, because moving an expensive convert operation after a filter which does not depend on its output can save unnecessary computation. For example, in the Real Estate Search workload, the fastest plan will apply the highly-selective text-based filters before it runs the expensive image processing step for testing whether a listing is ``modern and attractive." Importantly, the reordering takes into consideration any dependencies between conversion and filter operations to guarantee the equivalence of the reordered plans. At the moment, \system{} relies on the programmer to explicitly state dependencies between convert and filter operators. In the future, we hope add the capability for \system{} to deduce this information automatically, in a process akin to acquisitional query processing \cite{madden2003acqp}.

Two plans that are logically reordered are semantically equivalent. However, if implemented naively, a logical reordering might create a new set of physical LLM prompts that yield substantively different results.  Ensuring that the physical plan faithfully reflects the logical plan semantics is important and is described in \autoref{sec:physicalopt} below.



\subsection{Physical Optimizations}
\label{sec:physicalopt}
A number of recent AI optimizations, such reducing text generation latency via speculative inference~\cite{liu2023online, spector2023, leviathan2023fast, chen2023accelerating}, can naturally be deployed in many contexts, including chat, general LLM API services, and \system. But some optimizations are better or are only possible because our system's declarative programming framework permits  multiple low-level implementations that are semantically equivalent to the user's goal; such optimizations are unavailable when programming AI applications via a lower-level interface (such as direct prompting). We have implemented some of these optimizations in \system\ today, while others comprise future work.

\noindent {\bf Model Selection} --- which is implemented in our prototype --- is a simple but effective example of such an optimization. A high-level AI program comprises many tasks; for example, the code in \autoref{fig:enron-email-program} entails extraction of the {\tt Email} fields from unstructured text as well as filtering the emails by content. Thanks to its declarative language, \system\ can decompose the high-level program into smaller operations and choose the most appropriate model for each. It might be fine to use a cheap, small, fast model for easy operations, and only use the expensive model for harder ones. The idea is simple, but implementing it is not: because (1) a single program can comprise many operations, (2) the exact operation decomposition can change depending on other optimizations, (3) the difficulty of an operation can change over time, and (4) model quality can fluctuate as LLM services make updates. This optimization can easily become burdensome without \system's help.

\noindent {\bf Code Synthesis} --- which is implemented in our prototype --- involves generating synthesized code to perform specific operators dynamically. In certain real-world scenarios, tasks may not require deep semantic understanding and can be efficiently handled through synthesized code. By replacing calls to LLMs with calls to synthesized functions, we can generally reduce runtime and lower costs. The SEED system \cite{chen2024seed} used this approach to generate cost-optimized data curation pipelines with ensembles of synthesized code and LLM calls. In a similar vein, the {\sc Evaporate} system~\cite{arora2023language} used synthesized code, along with some weak supervision methods~\cite{Ratner_2017}, to replace some LLM operations. \system{} analyzes a set of sample inputs for a given conversion operation and then employs an LLM to generate a function that performs the necessary conversion. This approach can streamline processes where semantic depth is unnecessary, optimizing both performance and resource utilization.

\noindent {\bf Multi-data Prompt Marshaling} --- which is implemented in our prototype --- is the problem of deciding how user operators should be decomposed into prompts. Should data objects be processed in a row- or column-centric manner? For example, when running the code at the top of \autoref{fig:examples}, computing all of the {\tt ScientificPaper} fields in a single LLM call might allow us to process the input tokens just once, while obtaining multiple output values. But if there is a filter on the {\tt title} field, and if the {\tt citation} output field is very large, then a "column"-centric approach might be better.  Furthermore, different LLMs may be able to process more or fewer examples per invocation, depending on their context window and overall effectiveness.
Since the best decision will depend on the current state of the AI program and its input data, a human engineer attempting to implement this optimization manually would need to constantly reevaluate and possibly reimplement what goes into a particular prompt. 

\noindent {\bf Input Token Reduction} --- which is implemented in our prototype --- aims to delete portions of certain operators' input data while still obtaining high-quality results. For example, it should be possible to populate the {\tt title} and {\tt authors} fields of {\tt ScientificPaper} in \autoref{fig:examples} while ignoring almost all of the input text.  In document-processing use cases, this reduction in token size --- and thus reduction in financial cost and increase in execution speed --- can be very dramatic. By observing multiple naive examples of using {\tt convert()}  to transform a PDF to a {\tt ScientificPaper}, the declarative optimizer can run automatic "experiments" to determine which regions of the input are necessary. 
In some ways, this optimization is akin to a "micro-RAG" task, choosing salient excerpts from a source object. However, unlike a full RAG system, this optimization exists only for the life of the program execution. This is possible because \system{} can learn operator properties at the schema level. In the naive chat-processing use case, which lacks schemas and a visible workload of tasks from a single program, it is not clear how this method could be implemented. 

\noindent {\bf Output Token Reduction} aims to reduce the size of LLM generation outputs without changing the application-specific quality of these outputs. SplitWise \cite{patel2023splitwise} showed that LLM inference time is generally proportional to the size of the input and output tokens, with the output tokens having a greater impact when various LLM optimizations are activated. When \system{} is asked to find large excerpts from input objects, such as populating the {\tt citations} field of {\tt ScientificPaper}, the output token size can be very large. In these cases, it may be possible to reformulate the naive operation so it gives the same answer but with dramatically smaller output. For example, the system can insert artificial tokens into the input text and then ask the LLM to use these tokens to report which text block contains the target result.  In some early test cases, this approach can reduce output sizes from thousands of tokens down to just two.

\noindent {\bf Model Cascades} is a method that many researchers have explored in the context of image processing~\cite{10.14778/3137628.3137664, anderson2019physical, Cai2015LearningCC, Sun2013DeepCN, cascadeserve}. The core idea is to construct and exploit a series of models that accomplish the same goal, ranging from  fast/low-quality models on the low-end to expensive/high-quality models on the high-end. Operator processing starts on the inexpensive model. If the model can process the input with high confidence, the system uses its output; if not, the system gives the input to a more expensive model in the sequence. This approach is not implemented in \system\ yet, but it has been very successful in other related projects.

\noindent {\bf Knowledge Distillation} methods approximate the outputs of an expensive large-parameter model by using a smaller, cheaper (and often more limited) replica model~\cite{gu2024minillm, hinton2015distilling, Gou_2021}. Methods such as the {\em teacher / student} approach have produced small models that perform similarly to much larger ones, such as Alpaca and Vicuna~\cite{alpaca, vicuna2023}. In certain cases, it may be feasible for our system to perform knowledge distillation to obtain a small and fast model on a per-operator basis.

\noindent {\bf Workload-Aware Execution Management}, enabled by \system's ability to design entire workloads of model requests, should yield opportunities in model serving and resource management. If a distributed model service permits clients to give batch-oriented optimization hints --- such as whether a set of queries will use the same model, or to implement prefill prepacking~\cite{zhao2024prepacking} --- \system\ can exploit it.  Strategically co-scheduling LLM processing with similar prompts can optimize key-value (KV) cache reuse, substantially improving the KV cache hit rate during LLM inference \cite{liu2024optimizing,ye2024cascade}. Because the system can reformulate prompts and gather statistics about output token lengths, it should be able to design batches of model requests based on similar output token lengths, thereby minimizing waiting times for requests that finish early and boosting model request throughput.  


\subsection{Choosing An Optimization}
\label{sec:choose-opt}
In order to be valuable, the system must not just hypothesize a valuable set of optimized plans, it must actually choose one that yields concrete benefits.  \system\ follows steps (marked with circled numbers) described visually at the top of \autoref{fig:intro}, and algorithmically in Algorithm~\ref{algo:optimization}.

\begin{algorithm}
\caption{Optimized Plan Selection Algorithm}
\label{algo:optimization}
\begin{algorithmic}[1]
\REQUIRE $userCode$, $userPolicy$ \COMMENT{Step \circlednum{1}}
\STATE $logicalPlans$ = generateLogicalCandidates($userCode$) \COMMENT{Step \circlednum{2}}
\STATE $sentinelPhysicalPlans$ = getPhysicalPlans($logicalPlans$, $sentinel=True$) \COMMENT{Step \circlednum{3}}
\STATE
\STATE $performanceStatisics$ = \{\}
\FOR{0...NUM\_SAMPLES}
    \STATE $input$ = getSampledInput() 
    \STATE $stats$ = runAndComputeStatistics($sentinelPhysicalPlans$, $input$)  \COMMENT{Step \circlednum{4}}
    \STATE performanceStatistics.update($stats$) \COMMENT{Step \circlednum{5}}
\ENDFOR
\STATE
\STATE $physicalCandidates$ = getPhysicalPlans($logicalPlans$, $stats=performanceStatistics$)
\STATE $reducedCandidates$ = naiveElimination($physicalCandidates$)
\STATE $frontierCandidates$ = scoreAndEliminatePlans($reducedCandidates$, $performanceStatistics$)
\STATE
\RETURN chooseBestPlan($frontierCandidates$, $userPolicy$) \COMMENT{Step \circlednum{6}}
\end{algorithmic}
\end{algorithm}

The algorithm starts with a user program and an optimization goal (such as "minimize runtime subject to F1 > 0.5").  On line 1, it generates all possible logical plans consistent with the user program, as described in \autoref{sec:logical_opt}.  On line 2, it generates a small number of "sentinel" physical plans that are consistent with these logical plans. A sentinel plan is a simple hard-coded plan that should shed information on different selectivities and the accuracy of individual non-optimized operators. We currently generate three sentinel plans: one that uses GPT-3.5 for each convert and filter, one that uses MIXTRAL 8x7B, and one that uses GPT-4. (These are depicted in the sample-based statistics collection module at the top-right of \autoref{fig:intro}.)

On lines 4-8, the algorithm runs all of the sentinel plans on a small NUM\_SAMPLES number of inputs. The goal is to collect basic statistics about the behavior of some easy-to-understand plans.

On line 10, the system hypothesizes physical plans that embody optimizations for the logical plans.  This set can be quite large.  Our prototype generates 234, 10,140, and 1,950 physical plans for the Legal Discovery, Real Estate, and Medical Schema Matching programs, respectively.  On line 11, we use a small set of rules to eliminate plans that likely do not add to the set of useful runtime options.  On line 12, the algorithm uses statistics from lines 4-8 to score the expected runtime, financial cost, and quality of all the remaining physical plans; it then throws away any plans that are not on the Pareto frontier.

Finally, on line 14, the algorithm scores each frontier candidate in light of the user's optimization goal and returns the highest-scoring option for execution.  Estimating is done using well-known relational database planner methods, while estimating cost simply requires combining published service cost data with an accurate token consumption estimate.  Measuring quality is much harder, as described in \autoref{sec:quality}. The current system simply uses the champion model approach.

The algorithm is admittedly naive. For example, it does not have a rigorous termination criterion for the sampling phase on lines 5-8, and the sentinel plans are chosen arbitrarily. But as we will see in \autoref{sec:evaluation}, the algorithm effectively identifies plans that are close to the true optimal plan within the set generated by \system{}.

\section{Evaluation}
\label{sec:evaluation}

We evaluated \system{} on the three workloads motivated in \autoref{sec:intro}. We first describe our current prototype, a \system{} program for each workload, and our evaluation methodology. We then examine two experimental claims. First, \system\ can use
 its implemented physical optimizations (model selection, code synthesis, multi-data prompt marshaling, and input token reduction) to obtain physical plans that offer diverse performance trade-offs, some of which are more appealing than unoptimized naive plans. Second, the \system\ optimizer can predict plan properties quickly and accurately enough to choose a plan that fits user preferences better than unoptimized naive plans. Finally, we showcase \system{}'s ability to leverage system parallelism to minimize workload runtimes.

\subsection{Our Prototype}

We have implemented an early \system\ prototype in about 9,200 lines of Python code, which we have made available at \url{https://github.com/mitdbg/palimpzest}. It implements the operators in \autoref{fig:algebra}, and can run many programs, including the three test workloads described in \autoref{sec:intro} --- Legal Discovery, Real Estate Search, and Medical Schema Matching.

We have implemented four of the optimizations described in \autoref{sec:optimizations}: model selection, code synthesis, multi-data prompt marshaling, and input token reduction. We currently test the system using the {\tt gpt-3.5-turbo-0125}, {\tt gpt-4-0125-preview}, and {\tt gpt-4-vision-preview} OpenAI models~\cite{openaiapi} and the {\tt Mixtral-8x7B-Instruct-v0.1} model served by the Together.ai API~\cite{togetherai}. We also use the Modal online service~\cite{modalapi} for bulk non-AI function execution, such as parallel PDF processing and equation image extraction and conversion. We have tested the system with local model execution (via the Ollama~\cite{ollamawebsite} framework), but so far we have rarely found this to be an appealing option.

\system{} is implemented using the iterator model. Thus, plan execution proceeds one record at a time with each operator blocking until it receives the necessary input record(s) from its source operator(s). For clarity's sake, most of our experiments report simple single-threaded execution time, so we can better show the work saved by system optimizations.  However, many operations --- including the convert operator --- have parallel implementations that take advantage of parallelism offered by the underlying service or hardware, and we report a few parallel numbers.

\subsection{Evaluation Workloads}
We evaluated the model selection, code synthesis, and input token reduction optimizations described in \autoref{sec:optimizations} using the three workloads first discussed in ~\autoref{sec:intro}. Below, we provide details on the implementation and data used for each workload.

\begin{figure}
  \centering 
  \begin{minted}[xleftmargin=17pt,linenos,escapeinside=||,fontsize={\fontsize{8.5}{7.5}\selectfont}]{python}
import palimpzest as pz

class RealEstateListingFiles(|\textcolor{blue}{pz.Schema}|):
    """The source text and image data for a real estate listing."""
    listing = |\textcolor{blue}{pz.StringField}|(desc="The name of the listing")
    text_content = |\textcolor{blue}{pz.StringField}|(desc="The content of the listing's text description")
    image_contents = |\textcolor{blue}{pz.ListField}|(element_type=|\textcolor{blue}{pz.BytesField}|, desc="A list of the image contents")

class TextRealEstateListing(|\textcolor{blue}{RealEstateListingFiles}|):
    """Represents a real estate listing with specific fields extracted from its text."""
    address = |\textcolor{blue}{pz.StringField}|(desc="The address of the property")
    price = |\textcolor{blue}{pz.NumericField}|(desc="The listed price of the property")

class ImageRealEstateListing(|\textcolor{blue}{RealEstateListingFiles}|):
    """Represents a real estate listing with specific fields extracted from its images."""
    is_modern_and_attractive = |\textcolor{blue}{pz.BooleanField}|(desc="The home interior is modern and attractive")
    has_natural_sunlight = |\textcolor{blue}{pz.BooleanField}|(desc="The home interior has lots of natural sunlight")

def within_two_miles_of_mit(record):
    # filter based on record.address

def in_price_range(record):
    # filter based on record.price

# define logical plan
listings = |\textcolor{blue}{pz.Dataset}|(source="real-estate-eval", schema=|\textcolor{blue}{RealEstateListingFiles}|)
listings = listings.convert(|\textcolor{blue}{TextRealEstateListing}|, depends_on="text_content")
listings = listings.convert(|\textcolor{blue}{ImageRealEstateListing}|, depends_on="image_contents")
listings = listings.filter(
    "The interior is modern and attractive, and has lots of natural sunlight",
    depends_on=["is_modern_and_attractive", "has_natural_sunlight"]
)
listings = listings.filter(|\textcolor{blue}{within\_two\_miles\_of\_mit}|, depends_on="address")
listings = listings.filter(|\textcolor{blue}{in\_price\_range}|, depends_on="price")

# create and execute physical plans...
    \end{minted}
  \caption{The AI program written using \system{} for the Real Estate Search workload. 
  }
  \label{fig:real-estate-program}
\end{figure}

{\bf Legal Discovery.} We implemented this task using the source code shown in \autoref{fig:enron-email-program}.  The program loads the raw text files and stores their {\tt filename} and {\tt contents} in a {\tt pz.TextFile}. On line 9, the program converts the input data to the {\tt Email} schema, computing the {\tt sender} and {\tt subject} fields. Lines 10 and 11 filter the email by content. 

We created a test dataset of 1000 emails drawn from the Enron email collection~\cite{enron}. This dataset was curated and labeled by hand to contain 50 examples that discuss the management of fraudulent investment vehicles, with another 30 that contain fraud-related text but are not fraudulent in nature. The remainder were randomly chosen.  We registered this set of 1000 emails using the identifier ``enron-eval". The goal of the program is to recover only those emails that indicate genuine fraudulent activity.  For all the experiments on this workload, we evaluated the F1-score of the program's output using the manual labels described above. (Of course, during optimization and plan selection, \system\ did not have access to these labels and had to estimate the plans' quality in an unsupervised fashion.)

{\bf Real Estate Search.}  Our \system{} program for this workload is shown in \autoref{fig:real-estate-program}. It has the same rough form as the Legal Discovery task above, with a few key differences.  First, because the real estate listings are multimodal, they currently do not fit into a single core library schema such as {\tt pz.TextFile}. Thus, we first define a few utility schemas. {\tt RealEstateListingFiles} stores the raw image and text data for each listing; the data loading is performed by a simple user-defined datasource that we have omitted for brevity. We also define {\tt TextRealEstateListing} and {\tt ImageRealEstateListing} for the text and image attributes that we wish to compute~\footnote{At the moment, writing two schemas is necessary because schemas and convert operations have a one-to-one relationship in our system, but in the future we plan to support splitting a single schema conversion across multiple convert operations.}.  Second, on lines 33 and 34 the user provides user-defined functions for selection, instead of text strings.  Finally, several operations on lines 27-34 use the {\tt depends\_on} parameter (which was described in \autoref{sec:logical_opt}), in this case, allowing the system to skip image processing entirely.

We manually scraped 100 real estate listings from Boston and Cambridge, MA to obtain their natural language text descriptions and the first three deduplicated images from each listing.  We registered this dataset using the identifier ``real-estate-eval".  For each listing we manually labeled whether it satisfied the location, price, and "attractive and modern" and "natural sunlight" criteria, allowing us to compute an F1-score for any program output. Overall, 23 of the 100 data items satisfied all criteria.

{\bf Medical Schema Matching.} The \system\ program for this task is shown in \autoref{fig:medical-schema-matching-program}. This program aims at reproducing the output of the data harmonization pipeline performed by the authors of~\cite{li2023proteogenomic}.
The authors provide a single table produced as the result of collecting data from 11 different experimental studies. In this original study, all tables from the various studies were manually combined to create a comprehensive view of patient information related to tumor cases. This is a complex and time-consuming task, even for domain experts. According to the author contribution section, 66 out of 79 paper authors focused on the data curation task. However, with \system, we can implement the pipeline in approximately 30 lines of code. Lines 3-21 of the program reflect a detailed description of the target output schema, as described in the paper. This program is similar to the programs for Legal Discovery and Real Estate Search, but its unique feature is the use of the {\tt cardinality} parameter, which allows \system\ to yield multiple output objects from a single input object, i.e., multiple tables from a single spreadsheet, and multiple case data from the tables.

We curated a dataset of 11 spreadsheet files --- containing a total of 49 tables --- from the supplemental data provided in the original works~\cite{Cao_2021, Clark_2019, Dou_2020, Huang_2021, Li_2023, Gillette_2020, Krug_2020, McDermott_2020, Satpathy_2021, Vasaikar_2019,Wang_2021}. 
In Figure~\ref{fig:medical-schema-matching-program}, we refer to this dataset as ``medical-eval''. The first goal of the program is to filter, out of the 49 input tables, only those which contain information about patient case data. Once these tables are identified, the program aims at mapping each column in the source tables to a target column in a set of 15 attributes. This task is very challenging: not all target attributes can be found in all source tables, and neither there is guarantee that matching columns in the source tables have the same representation as the target columns.

For this workload, we measure accuracy using manually annotated ground truth data matching the columns from the source tables to those of the harmonized table provided by the authors of~\cite{li2023proteogenomic}.
We report the micro-average of the F1-score across all target attributes and paper studies.

\subsection{Optimizations Produce Plans with Diverse Performance Trade-offs}
\label{sec:opt-and-tradeoffs}

\begin{figure}[h]
    \centering
    \includegraphics[width=0.95\textwidth]{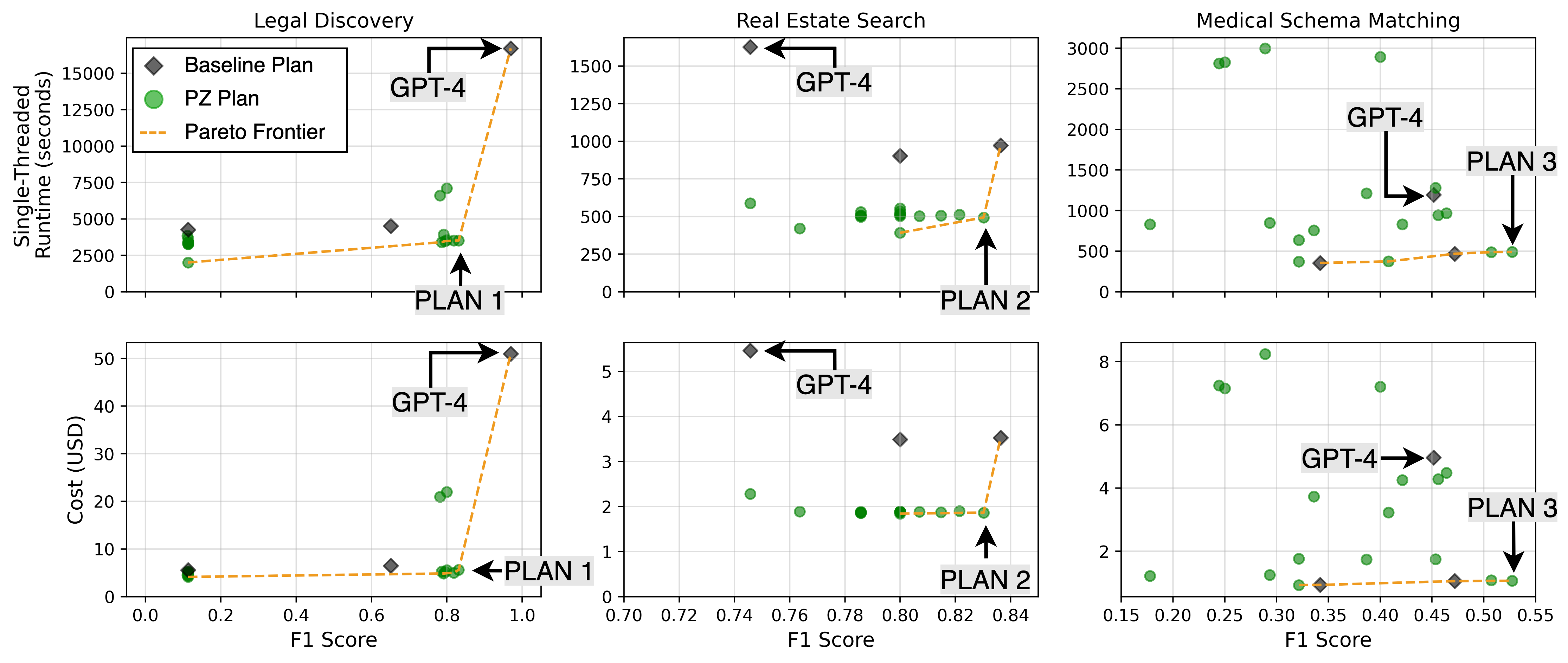}
    \caption{Performance of different plans on each workload (plans towards the bottom-right of each subplot are better). Naive plans are depicted with black diamonds and plans which \system{} creates with its physical and logical optimizations are shown with green circles. Across all workloads, \system{} produces plans on the Pareto frontiers of runtime vs. quality and cost vs. quality.}
    \label{fig:all-plans}
\end{figure}

Our first experimental claim is that \system\ can use its three optimization strategies to create a set of physical plans that offer appealing trade-offs regardless of the user policy.  To evaluate this claim, for each workload we ran \system{} up to (but not including) the final step in Algorithm~\ref{algo:optimization}. We then took the set of frontier plans, three baseline plans, and the top-$k$ plans from the {\it reducedCandidates} that were closest to the approximated Pareto frontier, such that we ultimately executed 20 plans in total. Our three baselines were the naive physical plans for each workload which used only GPT-4, GPT-3.5, or Mixtral-8x7B, respectively. The optimizer's compilation process took 2.6s, 13.1s, and 2.7s for Legal Discovery, Real Estate Search, and Medical Schema Matching, respectively.

Figure~\ref{fig:all-plans} shows the observed runtime, cost, and quality that came from executing all the aforementioned plans. Some plans exhibit similar performance, which makes them overlap in the figure, thereby presenting fewer than 20 distinct visual data points. Plans closer to the bottom-right of each subplot are better. Note that data points reflect observed values, not the system's estimated ones. During sample-based statistics collection, we used 5\% of the workload's total input size to run our sentinel plans, which gathered data to help estimate the cost of all plans. The one exception to this was the Medical Schema Matching workload, where we used 1 out of 11 inputs to gather sample data.

We found that \system{} is able to create useful plans at a number of different points in the trade-off space. On the Legal Discovery workload, \system{} was able to suggest physical plans (e.g. PLAN 1) that dominated the GPT-3.5 and Mixtral baselines, with an F1-score that is 7.3x and 1.3x better (respectively) at lower runtime and cost. Compared to the GPT-4 baseline, \system{} produced a cluster of plans near PLAN 1 that are $\sim$4.7x faster and $\sim$9.1x cheaper, while still achieving up to 85.7\% of the GPT-4 plan's F1-score.

The physical plan for PLAN 1 is shown in the Appendix. It achieved lower runtime and cost compared to the GPT-4 baseline plan by judiciously applying GPT-3.5 and Mixtral to the convert and filter operation(s). In particular, the plan did {\it not} use GPT-3.5 on the filter for fraudulent entities and did {\it not} use Mixtral on the filter for quotes from news articles. The models' poor performance on those filters, respectively, accounts for the worse performance of the baseline plans using only GPT-3.5 and only Mixtral. It is also worth noting that plans which used code synthesis for the convert operation provided speed-ups on that operation (with similar quality) relative to identical plans which used LLMs.  

On the Real Estate Search workload, where \system{} once again generated physical plans which offer desirable trade-offs relative to baseline plans. \system{} was able to produce physical plans (e.g. PLAN 2) that obtained 3.3x lower runtime, 2.9x lower cost, and up to 1.1x better F1-score than the GPT-4 baseline. These performance improvements are especially impressive considering the majority of the cost and runtime on this workload are dominated by calls to the vision model --- which cannot be optimized away using the methods in our current prototype. (In the future, we will explore options for visual processing optimizations.)

The physical plan for PLAN 2 is shown in the Appendix. It achieved improved performance relative to the GPT-4 baseline through the combination of two optimizations. First, the plan re-ordered the execution of the convert and filter operations such that the text-based operators were executed prior to image-based operators. This provided significant runtime and cost savings by avoiding calls to the GPT-4 vision model altogether. Second, the plan used input token reduction to trim the real-estate listing text by 50\%. This technique was particularly effective as the home address and listing price regularly appears at the top of the text for the listing.

Finally, on Medical Schema Matching --- our most challenging workload --- \system{} was able to produce plans (e.g. PLAN 3) which have $\sim2.4$x lower runtime, $\sim4.6$x lower cost, and up to 1.2x better F1-score than the GPT-4 baseline. This workload is especially challenging for \system{} because the code synthesis and token reduction optimizations are not very effective, but the system still identified plans that completely dominate the performance of a GPT-4 baseline.

The physical plan for PLAN 3 is shown in the Appendix. The plan used Mixtral for the filter and convert operations with a small amount token reduction (10\% of the input text trimmed). The use of Mixtral instead of GPT-4 provided significant speedups, and also provided better quality on this workload.  

We have shown that \system{} can generate plans like PLAN 1, 2, and 3 (plan details are shown in the Appendix), which provide users with compelling performance trade-offs. However, this does not automatically confirm that \system's optimizer will choose these plans during runtime. In the following section, we demonstrate that for a variety of policies, \system's cost optimizer does indeed identify and select such plans.




\subsection{Cost Optimizer Selects Plans with Significant Performance Improvements}
\label{sec:overallperformance}
\begin{figure}[h!]
    \centering
    \includegraphics[width=0.95\textwidth]{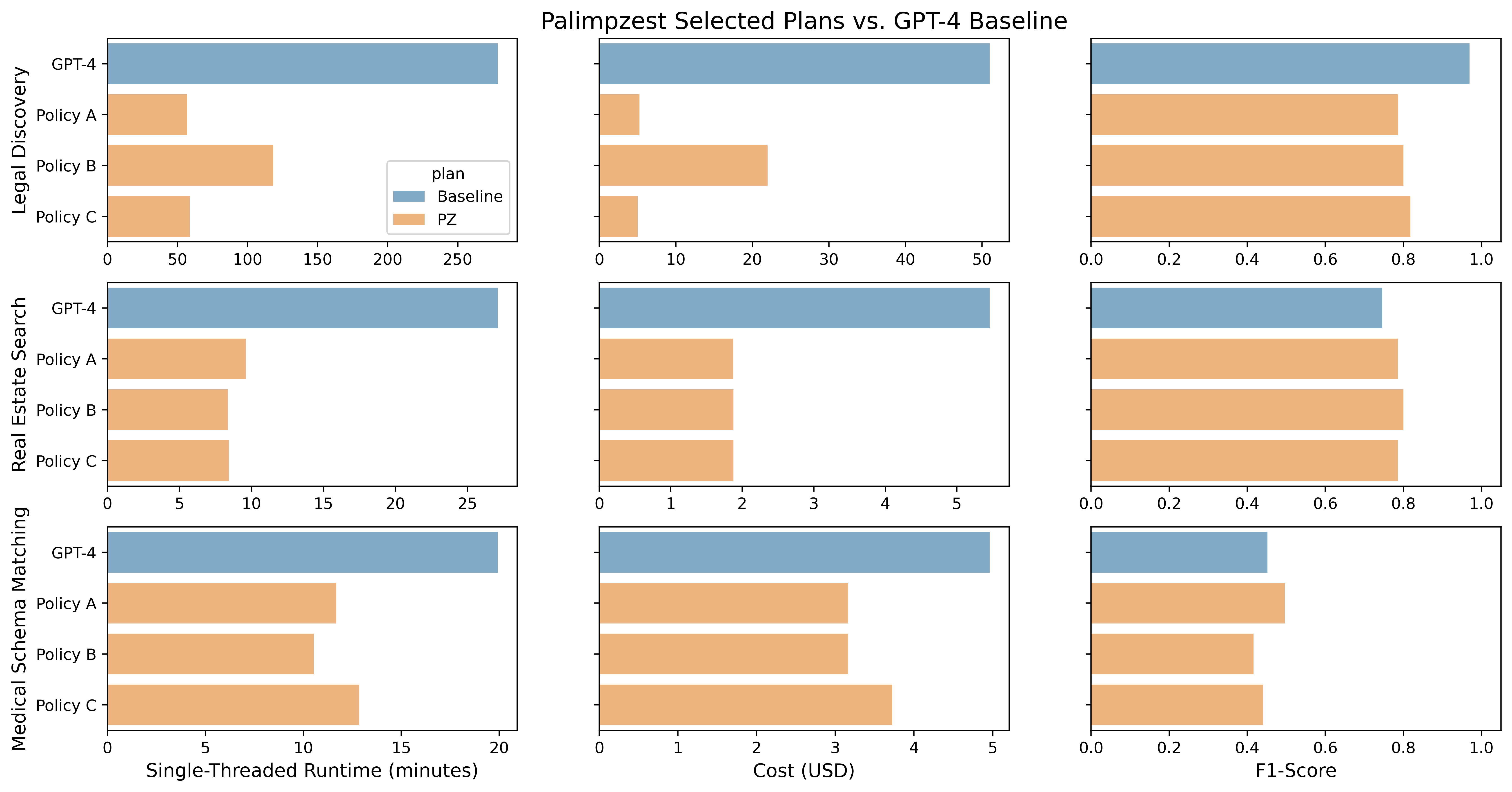}
    \caption{Comparing the performance of plan(s) selected by \system{} to the baseline GPT-4 plan for three different policies on each workload. {\bf Policy A} maximized F1-score at cost $<$ \$20, \$3, and \$2 for each workload (top-to-bottom); {\bf Policy B} maximized F1-score at runtime $<$ 167m, 10m, and 16.7m for each workload (top-to-bottom); and {\bf Policy C} minimized cost at F1-score $>$ 0.8, 0.8, and 0.4 for each workload (top-to-bottom). \system{} consistently finds plans which provide similar F1-scores and lower costs and runtimes than the GPT-4 baseline.}
    \label{fig:reopt}
\end{figure}

Our second experimental claim is that \system\ can identify plans that have better end-to-end runtime, cost, and quality than a naive plan that uses the same state-of-the-art language model for each operation. To evaluate this claim, we ran the system as described in Algorithm~\ref{algo:optimization} --- this time including the final step which chooses the best plan for a given policy. We ran three policies for each workload. {\bf Policy A} was to maximize quality at cost $<$ \$20.0, \$3.0, and \$2.0 for Legal Discovery, Real Estate Search, and Medical Schema Matching, respectively. {\bf Policy B} was to maximize quality at runtime $<$ 10,000s, 600s, and 1000s (for the same order of workloads). Finally, {\bf Policy C} was to minimize cost at an F1-score $>$ 0.8, 0.8, and 0.4 (for the same order of workloads). These fixed cost, quality, and runtime thresholds were set to be challenging, yet physically attainable based on our results in \autoref{sec:opt-and-tradeoffs}. (The full set of thresholds can be found in \autoref{tab:policy-results}.)

\autoref{fig:reopt} presents our results across three performance metrics, with each metric displayed in a separate column. The results are organized by workload, with each workload represented in a row of subplots. Within each subplot, results are further divided by the physical plan selected by the optimizer, with each plan occupying a separate row. We compared the plans chosen by \system{} to a baseline plan, which employs GPT-4 for all conversion and filtering operations.

Overall, we found that \system{} identifies plans in the space of physical candidates which (1) offer significant performance improvements over the GPT-4 baseline, (2) generally satisfy policy constraints (7 out of 9 satisfied), and (3) have speedups and cost savings which outweigh the overhead of collecting sample data. For example, on the Legal Discovery workload (first row in \autoref{fig:reopt}), the plan selected by \system{} for {\bf Policy A} achieved a runtime and financial cost that are 80.0\% and 89.7\% lower than the GPT-4 baseline, respectively, at an F1-score within 81.1\% of the baseline. The system achieved similar results for {\bf Policy C}, with slightly better quality (within 84.3\% of the baseline).


For the Real Estate Search workload (second row in \autoref{fig:reopt}), the plans chosen by \system{} achieved (on average) 67.5\% lower runtime, 65.7\% lower cost, and 6\% better F1-score than the baseline GPT-4 plan. Finally, on the Medical Schema Matching workload (third row in \autoref{fig:reopt}), \system{} was able to identify plans that provide up to 47.2\% and 36.3\% lower runtime and cost, respectively, at comparable F1-scores to the GPT-4 baseline.

\begin{table}
    \centering
    \caption{\system{} satisfied the policy constraint on 7 out of 9 plans it executed in \autoref{sec:overallperformance}. Constraint thresholds were chosen to be challenging, but realistic based on our results in \autoref{sec:opt-and-tradeoffs}.}
    \renewcommand{\arraystretch}{1.3}
    \centering
    \begin{tabular}{|l|l|l|c|}
    \hline
    {\bf Policy Name} &{\bf Policy Detail} & {\bf Workload } & {\bf Actual v. Constraint} \\
    \hline
    Policy A & Max F1 @ Cost $<\$20$ & Legal Discovery & {\bf \$5.27$<$\$20} \\
    \hline
    Policy A & Max F1 @ Cost $<\$3$ & Real Estate Search & {\bf \$1.87$<$\$3} \\
    \hline
    Policy A & Max F1 @ Cost $<\$2$ & Medical Schema Matching & \$3.16$\nless$\$2 \\
    \hline
    Policy B & Max F1 @ Runtime $<$ 10000s & Legal Discovery & {\bf 7102s$<$10000s} \\
    \hline
    Policy B & Max F1 @ Runtime $<$ 600s & Real Estate Search & {\bf 502s$<$600s} \\
    \hline
    Policy B & Max F1 @ Runtime $<$ 1000s & Medical Schema Matching & {\bf 632s$<$1000s} \\
    \hline
    Policy C & Min Cost @ F1 $>$ 0.80 & Legal Discovery & {\bf 0.82$>$0.80} \\
    \hline
    Policy C & Min Cost @ F1 $>$ 0.80 & Real Estate Search & 0.79$\ngtr$0.80 \\
    \hline
    Policy C & Min Cost @ F1 $>$ 0.40 & Medical Schema Matching & {\bf 0.44$>$0.40} \\
    \hline
    \end{tabular}
    \label{tab:policy-results}
\end{table}



\subsection{Minimizing Runtime with Parallel Operators}

\begin{table*}[h!]
    \caption{Runtime speedup from executing \system{} plans with parallel convert and filter operators. We compare these to the single-threaded GPT-4 baselines for each workload. All \system{} plans were selected by the system under {\bf Policy A}.}
    \begin{subfigure}{\textwidth}
        \caption{Legal Discovery}
        \vspace{-4pt}
        \renewcommand{\arraystretch}{1.3}
        \centering
        \begin{tabular}{|c|c|c|c|}
        \hline
        {\bf Plan } & {\bf Runtime (s)} & {\bf Cost (\$)} & {\bf F1}\\
        \hline
        Single-Threaded Baseline (GPT-4) & 16,712 & 51.0 & 0.97 \\
        \hline
        Palimpzest & 185 (1.1\%) & 5.60 (11.0\%) & 0.81 (83.5\%) \\
        \hline
        \end{tabular}
        \label{tab:speedup-legal-discovery}
    \end{subfigure}
    \begin{subfigure}{\textwidth}
        \caption{Real Estate Search}
        \vspace{-4pt}
        \renewcommand{\arraystretch}{1.3}
        \centering
        \begin{tabular}{|c|c|c|c|}
        \hline
        {\bf Plan } & {\bf Runtime (s)} & {\bf Cost (\$)} & {\bf F1}\\
        \hline
        Single-Threaded Baseline (GPT-4) & 1,626 & 5.46 & 0.75 \\
        \hline
        Palimpzest & 80.9 (5.0\%) & 1.86 (34.1\%) & 0.80 (107\%) \\
        \hline
        \end{tabular}
        \label{tab:speedup-real-estate}
    \end{subfigure}
    \begin{subfigure}{\textwidth}
        \caption{Medical Schema Matching}
        \vspace{-4pt}
        \renewcommand{\arraystretch}{1.3}
        \centering
        \begin{tabular}{|c|c|c|c|}
        \hline
        {\bf Plan } & {\bf Runtime (s)} & {\bf Cost (\$)} & {\bf F1}\\
        \hline
        Single-Threaded Baseline (GPT-4) & 1,195 & 4.96 & 0.45 \\
        \hline
        Palimpzest & 215 (18.0\%) & 3.36 (67.7\%) & 0.46 (102\%) \\
        \hline
        \end{tabular}
        \label{table:speedup-medical}
    \end{subfigure}
    \label{tab:speedup}
\end{table*}



For our final evaluation, we ran \system{} with parallel implementations of the convert and filter operations to demonstrate the system's ability to achieve large runtime speedups --- with competitive costs and F1-scores --- relative to a single-threaded baseline plan. For each convert and filter operator we used 32 workers to execute operations in parallel. We did not pipeline operations, i.e. every record needed to be processed in a given convert or filter operation before the next operation in the plan could commence (but this is an optimization we could implement in the future). For each workload we ran \system{} and asked the optimizer to select the optimal plan according to {\bf Policy A}.

The results of our evaluation are shown in \autoref{tab:speedup}. We can see that on Legal Discovery, \system{} achieved a {\bf 90.3x speedup at 9.1x lower cost while obtaining an F1-score within 83.5\% of the GPT-4 baseline}. On Real Estate Search and Medical Schema matching, the optimized plan dominated the GPT-4 baseline on all metrics. {\bf These plans achieve better F1-scores than their baselines, and do so with speedups of 20.0x and 5.6x as well as cost savings of 2.9x and 1.5x, respectively.}

We do not use any exotic algorithms, and of course it is straightforward to run model prompts in parallel.  \system's abstractions simply allow the system to obtain these speedups with no additional work by the user.

\section{Related Work}
\label{sec:relatedwork}
There is a large and growing literature in programming frameworks for LLMs, foundation models, and other AI artifacts.

Many current frameworks are focused on prompt management. 
LangChain~\cite{langchain} and LlamaIndex~\cite{Liu_LlamaIndex_2022} are popular libraries for building LLM applications, with special support for mechanical issues surrounding AI model use: managing prompt templates, handling user-provided labeled examples, managing document tokenization, and so on. They have no higher-level task representation, although there is support for standard use cases like chatbots and RAG.

DSPy~\cite{dspy} focuses on creating high-quality prompts. It takes a high-level description of a user's LLM goal and yields a set of concrete prompts (or even LLM operations like fine-tuning) to implement the user's goal. Like \system, DSPy aims to abstract some details of the AI programming process. But DSPy's goals are different: it focuses entirely on improving prompt quality, ignores most performance and cost issues, and asks the programmer to make decisions about its machine learning algorithm.  We have used DSPy in some cases when compiling individual tasks inside \system.

Several other projects --- such as Outlines~\cite{willard2023efficient}, Guidance~\cite{guidance_ai_2023}, and RELM~\cite{kuchnik2023validating} --- attempt to control how LLMs emit data. These projects are especially useful when producing data that is intended for machine consumption, such as producing a JSON record that conforms to a particular schema, or emitting classification labels drawn from a target vocabulary. Outlines and RELM offer runtime optimizations that avoid inference costs when the syntax fully determines the output. These are useful systems but are also fairly narrow and do not address our target workload. 

SGLang~\cite{zheng2023efficiently} is an ambitious project that combines a number of features present in the above systems, such as prompt templating methods and constrained outputs. It also tries to offer some performance features not present in the above systems, such as parallel execution and batching. It even offers management of multimodal data. However, the Structured Generation Language is still fairly low-level and asks developers to make many decisions manually, such as the direct text of a prompt, the desired elements that comprise a batch of  prompts; or the degree of parallelism. It might make sense for a future version of \system\ to use SGLang as a compiled runtime description language.

The SkyPilot system~\cite{Yang2023SkyPilotAI} is similar to \system\ in its emphasis on cost reduction of ML workloads, but is primarily concerned with efficiently deploying coarse DAG of tasks across broadly-similar cloud platforms.  It offers a broker that stands between a user's workload and advertised services from competing cloud services. It aims for an ideal assignment of work to resources, rather than formulating efficient program-specific implementations.

FrugalGPT~\cite{chen2023frugalgpt} is a high-level library meant to be deployed on top of LLM access packages such as OpenAI~\cite{openaiapi}. It offers a number of valuable optimizations, some of which resemble the optimizations in \autoref{sec:optimizations}. The FrugalGPT vision of appealing AI optimizations, combined with some trade-offs, is entirely consistent with \system's goal and approach, and the observed improvements in runtime and cost are strong. However, it is unclear how the user describes tasks to the FrugalGPT's optimizer, so the scope for future optimization is also unclear. Moreover, it is not clear whether FrugalGPT optimizations can be implemented in the context of a broader AI program that includes not just LLM processing but also non-LLM AI components and conventional data processing elements.

AutoGen~\cite{wu2023autogen} is an interesting development framework for "LLM applications," but with a domain focus that is dramatically different from the one we pursue. This system imagines LLM applications as conversations among agents, which might be humans, LLMs, or tools. The AutoGen developer aims to design the conversation and control flow that takes place among these agents. This vision of LLM applications is driven by a fundamentally different model from our own, which is based more in the data processing tradition.

Urban and Binnig proposed "Language-Model-Driven Query Planning" in the Caesura system~\cite{urban2023caesura}. The goal is to allow users to describe a data processing task in natural language, which yields an executable query plan, in particular one that can operate on multimodal data. This work is similar to \system\ in that it yields a query plan that combines traditional relational operators with model-driven ones. However, \system\ makes no attempt at all to build the logical query plan from natural language; we expect a human programmer to provide the core processing logic. The published Caesura system does not perform any plan optimization, although the authors describe this as an area for future work.

ZenDB \cite{lin2024towards} is a document analytics system that enhances query performance by extracting semantic hierarchical structures from templatized documents and integrating a query engine that leverages these structures to efficiently and accurately execute SQL queries on document collections. ZenDB employs optimization techniques such as predicate reordering, pushdown, and projection pull-up to refine the execution of user-specified SQL queries. This optimization is facilitated through a summary-based tree search for each document, aiming to minimize cost and latency while maximizing accuracy. Although ZenDB and similar systems share optimization objectives with \system, their efforts primarily concentrate on logical optimization and are predominantly focused on managing text data. In contrast, \system\ utilizes multiple strategies to optimize broader aspects of query execution.

The Evaporate system proposed by Arora et al.\ in~\cite{arora2023language} shares some intuitions of \system{} regarding the trade-off of cost/quality in LLM workloads using code generation.
However, while Evaporate is designed specifically for an information extraction use case and employs only static prompts and code generation, \system{} provides a more general approach to express a wider class of ML workloads, with several optimization strategies alongside code generation, e.g., model selection and token reduction.
As such, we believe the more advanced weak-supervision code generation strategy proposed in~\cite{arora2023language} could be integrated as one of the optimizations within the \system{} framework.



\system{} has some commonalities with  CrowdDB~\cite{crowddb}, Qurk~\cite{qurk}, and Deco~\cite{deco};  systems that aim to provide a declarative approach to create data processing pipelines with crowd-sourcing operations to answer queries that cannot ``be answered by machines only'' \cite{crowddb}. 
For example, ~\cite{crowddb} automatically optimized data processing pipelines with  crowd-sourcing operations for entity matching or (open) data retrieval, whereas ~\cite{crowdcount} introduced and optimized crowd-sourcing operators to extract data from images. 
It turns out that --- with the power of LLMs --- some of the tasks no longer need humans and can be answered by machines only.
Instead of using crowd-sourcing operators which give humans tasks to do, LLMs can perform tasks like extracting data from text, images, etc. much faster and often with better quality. 
However, any task done by an LLM still has a quality/time/cost trade-off and, similar to crowd-sourcing tasks, the quality increases the more one is willing to pay per task (e.g., by using a large vs small model, single vs multiple invocations, etc.). 
However, while this trade-off creates some similarities between \system{} and crowd-sourcing systems, \system{}'s use of LLMs provides more degrees of freedom (e.g., the ability to generate code in seconds), creates new sets of challenges (e.g., token optimization), and does not have to deal with ``lazy'' workers, which was one of the key challenges of declarative crowd-based systems.

\section{Conclusion}
\label{sec:conclusion}
\system{} enables users to program AI workloads in a declarative language, which it can then optimize in an efficient manner. As a result, users can focus on the high-level logic of their applications without getting bogged down in the intricacies of underlying AI models and optimization strategies. \system{} integrates logical and physical planning, in a best-effort attempt to execute each program as efficiently as possible. Further, \system's optimizer exploits sample execution data, allowing it to optimize and trade-off runtime performance, financial cost, and quality. Together, these features make \system{} a useful tool for developers and organizations aiming to take advantage of the full potential of modern AI capabilities in an efficient and affordable manner. \system's declarative language, combined with its automatic planning and optimization capabilities, presents a new opportunity for the development of semantic analytics applications.

\bibliographystyle{plain}
\bibliography{references}  

\section*{Appendix}
\label{sec:appendix}
\subsection*{Evaluation Plan Details}
In \autoref{sec:opt-and-tradeoffs} we highlight three plans (PLAN 1, 2, and 3), which provide useful performance trade-offs relative to baseline plans. The implementation details for each of these plans are described below.

\textbf{PLAN 1:} uses Mixtral-8x7B to convert each \texttt{TextFile} to the \texttt{Email} schema. It then uses Mixtral-8x7B again to apply the filter which checks whether or not the email refers to a fraudulent investment scheme. Finally, this plan uses gpt-3.5-turbo to determine whether or not the email is referring to a news article or is written by someone outside of Enron. A depiction of the plan is shown below:

\lstset{style=mystyle}
\begin{lstlisting}
 0. MarshalAndScanDataOp -> File 

 1. File -> InduceFromCandidateOp -> TextFile 
    Using hardcoded function
    (contents,filena...) -> (contents,filena...)

 2. TextFile -> InduceFromCandidateOp -> Email 
    Using Model.MIXTRAL
    Token budget: 1.0
    Query strategy: QueryStrategy.BONDED_WITH_FALLBACK
    (contents,filena...) -> (contents,filena...)

 3. Email -> FilterCandidateOp -> Email 
    Using Model.MIXTRAL
    Filter: "The email refers to a fraudulent scheme (i.e., "Raptor", ...)"
    (contents,filena...) -> (contents,filena...)

 4. Email -> FilterCandidateOp -> Email 
    Using Model.GPT_3_5
    Filter: "The email is not quoting from a news article..."
    (contents,filena...) -> (contents,filena...)
\end{lstlisting}

\textbf{PLAN 2:} uses gpt-3.5-turbo (with a token budget allowing it to process up to 50\% of the listing text) to convert each \texttt{RealEstateListingFiles} to the \texttt{TextRealEstateListing} schema. It then applies the UDF filters which check whether or not the listing is within two miles of MIT and in the user's price range. The plan then uses the gpt-4-vision-preview model to convert the schema to an \texttt{ImageRealEstateListing} (extracting the ``modern and attractive" and ``natural sunglight" attributes), before finally applying gpt-3.5-turbo to filter based on these attributes. A depiction of the plan is shown below:

\begin{lstlisting}
 0. MarshalAndScanDataOp -> RealEstateListingFiles 

 1. RealEstateListingFiles -> InduceFromCandidateOp -> TextRealEstateListing 
    Using Model.GPT_3_5
    Token budget: 0.5
    Query strategy: QueryStrategy.BONDED_WITH_FALLBACK
    (image_contents,...) -> (address,image_c...)

 2. TextRealEstateListing -> FilterCandidateOp -> TextRealEstateListing 
    Using None
    Filter: "<function get_logical_tree.<locals>.within_two_miles_of_mit"
    (address,image_c...) -> (address,image_c...)

 3. TextRealEstateListing -> FilterCandidateOp -> TextRealEstateListing 
    Using None
    Filter: "<function get_logical_tree.<locals>.in_price_range"
    (address,image_c...) -> (address,image_c...)

 4. TextRealEstateListing -> InduceFromCandidateOp -> ImageRealEstateListing 
    Using Model.GPT_4V
    Token budget: 1.0
    Query strategy: QueryStrategy.BONDED_WITH_FALLBACK
    (address,image_c...) -> (has_natural_sun...)

 5. ImageRealEstateListing -> FilterCandidateOp -> ImageRealEstateListing 
    Using Model.GPT_3_5
    Filter: "The interior is modern and attractive, and has lots of natural sunlight"
    (has_natural_sun...) -> (has_natural_sun...)
\end{lstlisting}

\textbf{PLAN 3:} uses Mixtral-8x7B to filter for the subset of tables which contain patient age information. It then uses Mixtral-8x7B once again (with a token budget allowing it to process up to 90\% of the input table data) to convert each input \texttt{Table} to the desired output \texttt{CaseData} schema. A depiction of the plan is shown below:

\begin{lstlisting}
 0. MarshalAndScanDataOp -> File 

 1. File -> InduceFromCandidateOp -> XLSFile 
    Using hardcoded function
    (contents,filena...) -> (contents,filena...)

 2. XLSFile -> InduceFromCandidateOp -> Table 
    Using hardcoded function
    (contents,filena...) -> (filename,header...)

 3. Table -> FilterCandidateOp -> Table 
    Using Model.MIXTRAL
    Filter: "The rows of the table contain the patient age"
    (filename,header...) -> (filename,header...)

 4. Table -> InduceFromCandidateOp -> CaseData 
    Using Model.MIXTRAL
    Token budget: 0.9
    Query strategy: QueryStrategy.BONDED_WITH_FALLBACK
    (filename,header...) -> (age_at_diagnosi...)
\end{lstlisting}

\begin{figure*}[h!]
    \begin{subfigure}[t]{0.48\textwidth}
        \includegraphics[width=\textwidth]{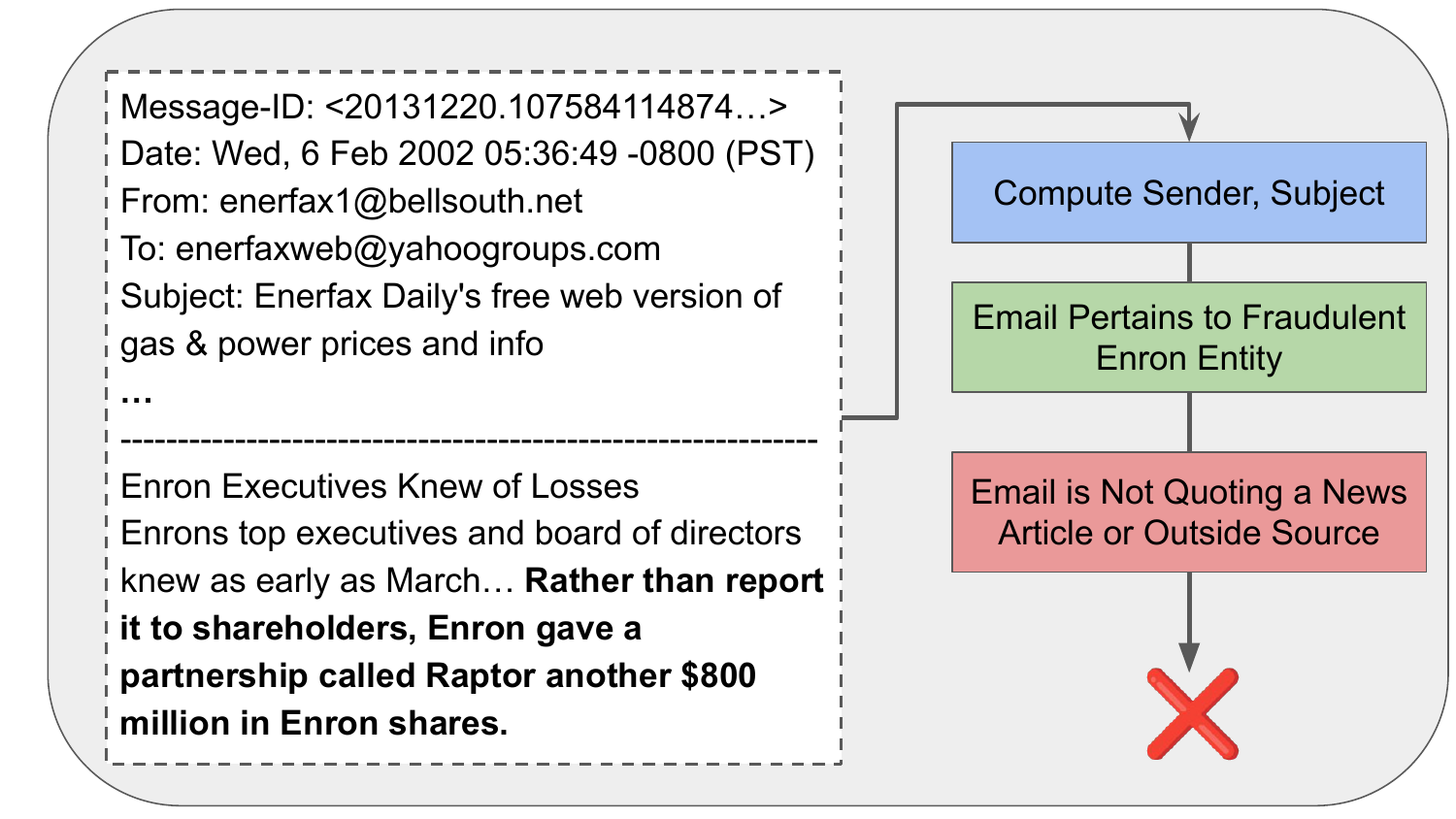}
        \caption{Example of a negative entry in the Legal Discovery workload. This email does not meet the criteria set by the user's search because the email is quoting from a news article.}
    \label{fig:enron-incorrect}
    \end{subfigure}
    \hfill
    \begin{subfigure}[t]{0.48\textwidth}
        \includegraphics[width=\textwidth]{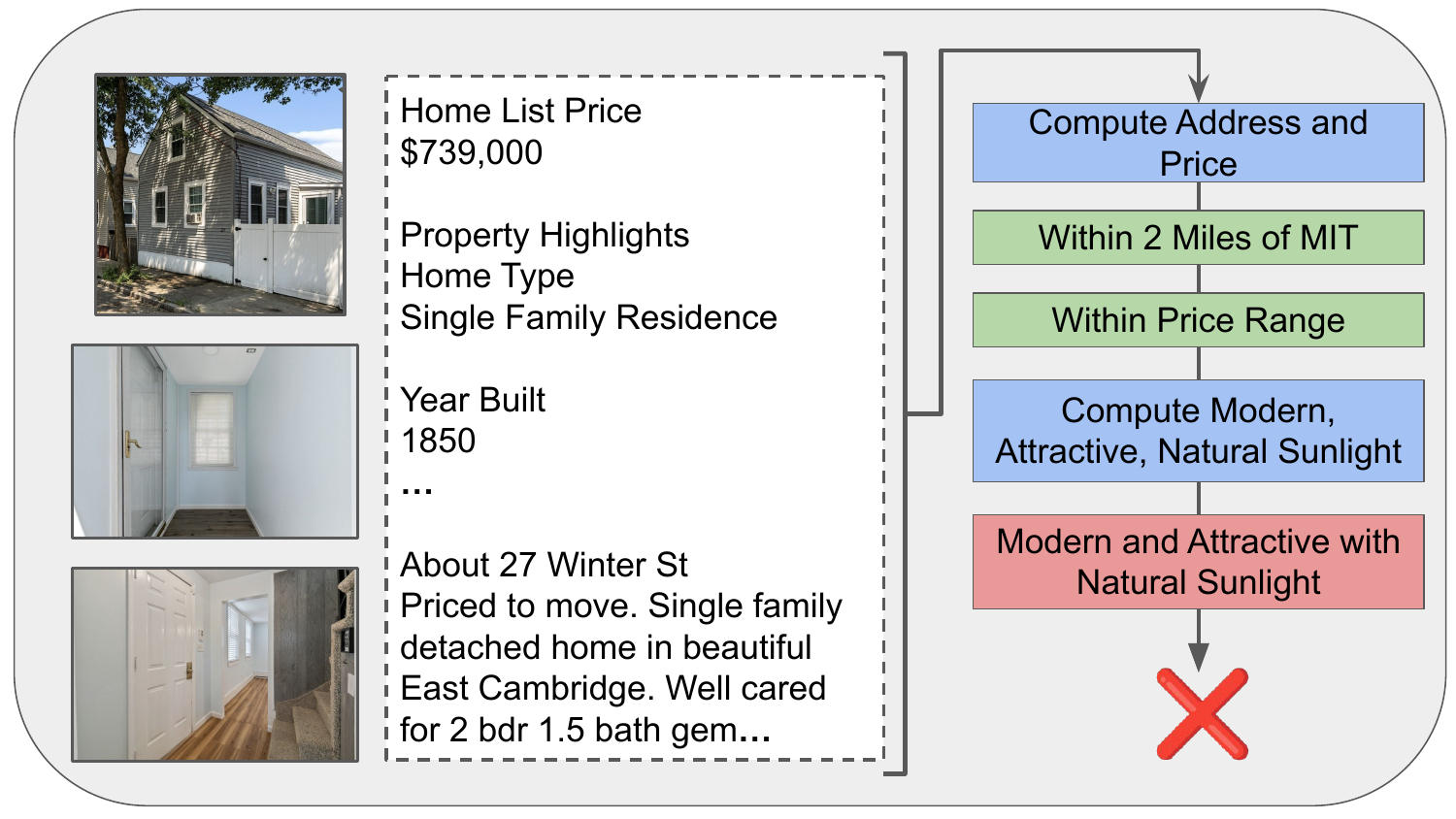}
        \caption{Example of a negative entry in the Real Estate Search workload. This house does not meet the criteria set by the user’s search because the house is not considered to be modern and attractive.}
    \label{fig:real-estate-search-incorrect}
    \end{subfigure}
    \caption{Negative examples in the Legal Discovery and Real Estate Search workloads.}
    \label{fig:negative-examples}
\end{figure*}


\begin{figure}
  \centering 
  \begin{minted}[xleftmargin=17pt,linenos,escapeinside=||,fontsize={\fontsize{8.5}{7.5}\selectfont}]{python}
import palimpzest as pz

class CaseData(|\textcolor{blue}{pz.Schema}|):
    """An individual row extracted from a table containing medical study data."""
    case_submitter_id = |\textcolor{blue}{pz.StringField}|(desc="The ID of the case")
    age_at_diagnosis = |\textcolor{blue}{pz.NumericField}|(desc="The age of the patient...", required=False)
    race = |\textcolor{blue}{pz.StringField}|(desc="An arbitrary classification of a...", required=False)
    ethnicity = |\textcolor{blue}{pz.StringField}|(desc="Whether an individual...", required=False)
    gender = |\textcolor{blue}{pz.StringField}|(desc="Text designations that identify gender.", required=False)
    vital_status = |\textcolor{blue}{pz.StringField}|(desc="The vital status of the patient", required=False)
    ajcc_pathologic_t = |\textcolor{blue}{pz.StringField}|(desc="The AJCC pathologic T", required=False)
    ajcc_pathologic_n = |\textcolor{blue}{pz.StringField}|(desc="The AJCC pathologic N", required=False)
    ajcc_pathologic_stage = |\textcolor{blue}{pz.StringField}|(desc="The AJCC pathologic stage", required=False)
    tumor_grade = |\textcolor{blue}{pz.StringField}|(desc="The tumor grade", required=False)
    tumor_focality = |\textcolor{blue}{pz.StringField}|(desc="The tumor focality", required=False)
    tumor_largest_dimension_diameter = |\textcolor{blue}{pz.NumericField}|(desc="The largest...", required=False)
    primary_diagnosis = |\textcolor{blue}{pz.StringField}|(desc="The primary diagnosis", required=False)
    morphology = |\textcolor{blue}{pz.StringField}|(desc="The morphology", required=False)
    tissue_or_organ_of_origin = |\textcolor{blue}{pz.StringField}|(desc="The tissue or organ...", required=False)
    filename = |\textcolor{blue}{pz.StringField}|(desc="The name of the file...", required=False)
    study = |\textcolor{blue}{pz.StringField}|(desc="The last name of the author of the study...", required=False)


# define logical plan
xls = |\textcolor{blue}{pz.Dataset}|("medical-eval", schema=|\textcolor{blue}{pz.XLSFile}|)
patient_tables = xls.convert(
    |\textcolor{blue}{pz.Table}|, desc="All tables in the file", cardinality="oneToMany"
)
patient_tables = patient_tables.filter("The rows of the table contain the patient age")
case_data = patient_tables.convert(
    CaseData, desc="The patient data in the table", cardinality="oneToMany"
)

# create and execute physical plans...
    \end{minted}
  \caption{The AI program written using \system{} for the Medical Schema Matching workload. 
  }
  \label{fig:medical-schema-matching-program}
\end{figure}






\end{document}